\documentclass[10pt]{iopart}

\usepackage{iopams}
\usepackage{url,hyperref,lineno, booktabs, natbib}
\usepackage[onehalfspacing]{setspace}
\usepackage{graphicx}
\usepackage{algorithm, algpseudocode}
\usepackage{mdframed}

\begin{document}

\title[Preprint. Under review.]{Unsupervised End-to-End Training with a Self-Defined Target}

\author{Dongshu Liu\,$^{1,*}$, Jérémie Laydevant\,$^{1}$, Adrien Pontlevy\,$^{1}$, Damien Querlioz\,$^{2}$ and Julie Grollier\,$^{1,*}$}

\address{$^{1}$Laboratoire Albert Fert, CNRS, Thales, Université Paris-Saclay, Palaiseau, France\\
$^{2}$Université Paris-Saclay, CNRS, Centre de Nanosciences et de Nanotechnologies, Palaiseau, France}

\vspace{10pt}
\begin{indented}
\item[]* Corresponding author
\end{indented}
\ead{dongshu.liu@cnrs.fr, julie.grollier@cnrs-thales.fr}

\begin{abstract}

Designing algorithms for versatile AI hardware that can learn on the edge using both labeled and unlabeled data is challenging. Deep end-to-end training methods incorporating phases of self-supervised and supervised learning are accurate and adaptable to input data but self-supervised learning requires even more computational and memory resources than supervised learning, too high for current embedded hardware. Conversely, unsupervised layer-by-layer training, such as Hebbian learning, is more compatible with existing hardware but doesn't integrate well with supervised learning. 
To address this, we propose a method enabling networks or hardware designed for end-to-end supervised learning to also perform high-performance unsupervised learning by adding two simple elements to the output layer: Winner-Take-All (WTA) selectivity and homeostasis regularization. These mechanisms introduce a 'self-defined target' for unlabeled data, allowing purely unsupervised training for both fully-connected and convolutional layers using backpropagation or equilibrium propagation on datasets like MNIST (up to 99.2\%), Fashion-MNIST (up to 90.3\%), and SVHN (up to 81.5\%).
We extend this method to semi-supervised learning, adjusting targets based on data type, achieving 96.6\% accuracy with only 600 labeled MNIST samples in a multi-layer perceptron. Our results show that this approach can effectively enable networks and hardware initially dedicated to supervised learning to also perform unsupervised learning, adapting to varying availability of labeled data.

\end{abstract}

\vspace{2pc}

\section{Introduction}

Autonomous AI devices operating at the edge would be much more adaptable and useful if they could learn continuously, making use of both precise but usually scarce labeled data as well as widely available unlabeled data \citep{Xiao2022memristive, Barhoush2022Semi, shakarami_survey_2020}. But for this, they need to be able to perform both supervised and unsupervised learning.

\begin{figure}[ht]
    \centering
    \includegraphics[width=1\textwidth]{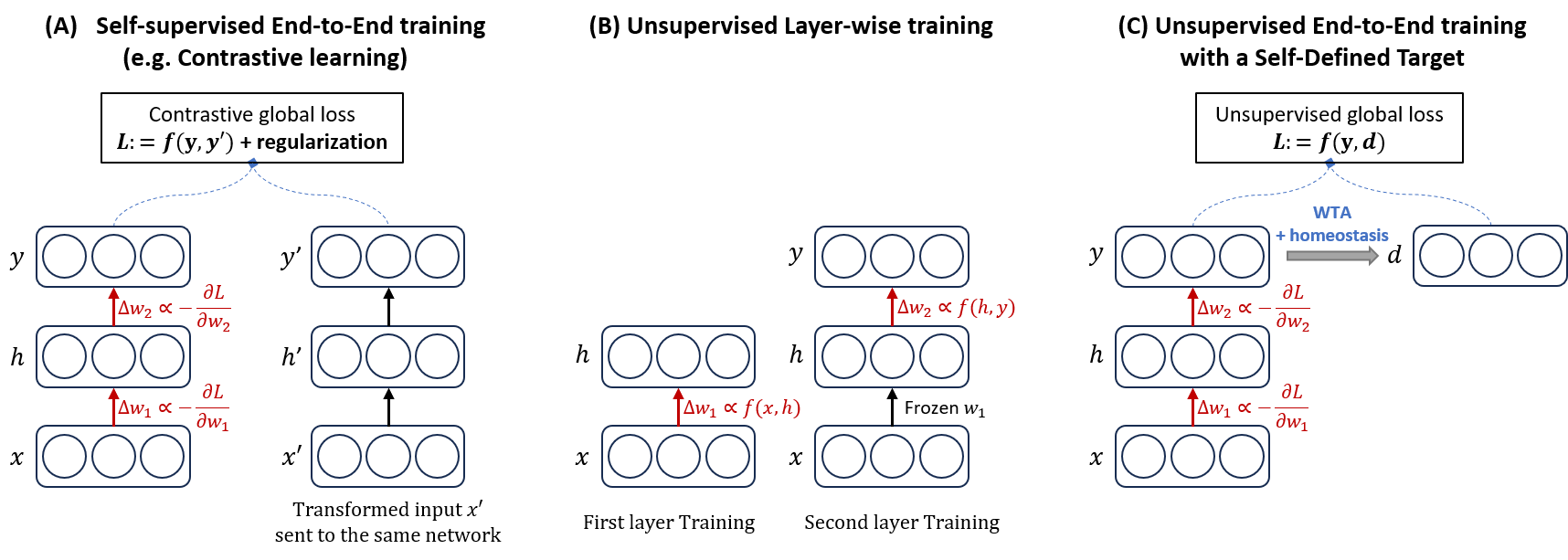}
    \footnotesize
    \caption{\label{fig:unsupervised-end2end}
(A) Self-supervised End-to-end training methods, here illustrated with a contrastive global loss as training objective. (B) Layer-wise training approaches typically used for unsupervised learning algorithms with local learning rules. (C) Our approach to unsupervised end-to-end using an unsupervised global loss defined at the network output.}
\end{figure}

State-of-the-art (SOTA) deep learning methods for unsupervised learning predominantly rely on self-supervised representation learning methods \citep{zbonstar2021barlow, bardes2022vicreg, siddiqui2022localssl, Balestriero2023ACO}, where all layers are trained together through backpropagation in the ``end-to-end'' manner schematized in Figure \ref{fig:unsupervised-end2end}(A). These methods can achieve high accuracy but necessitate gathering statistics (such as variance and covariance) of model outputs across mini-batches, demanding substantial computational and memory resources. 

These requirements cannot be satisfied in the case of edge computing. Typically, edge devices have restricted computational power, limited memory capacity, and stringent energy consumption requirements \citep{cao2020overview, chen2019DeepL}. Therefore, algorithms designed for edge computing must be resource-efficient to optimize power usage.

In parallel, unsupervised learning for edge AI hardware has explored bio-inspired approaches, 
such as Hebbian Learning \citep{do1949organization, azghadi2015programmable} or Spike-Timing-Dependent Plasticity (STDP) \citep{bi1998synaptic, bichler2012extraction, jo2010nanoscale, ishii2019chip, indiveri2006vlsi}, which circumvent the need for extensive computational resources. These methods enable learning to be both local and ``greedy'', meaning it occurs independently and sequentially at each layer as illustrated in Figure \ref{fig:unsupervised-end2end}(B) \citep{bengio2006greedy}. Recent works have shown impressive (and often thought impossible) results on ``Hebbian Deep Learning'' \citep{miconi2021hebbian, moraitis2022softhebb}, reaching high accuracy on complex tasks \citep{journe2022hebbian}. 
However, the absence of a global learning objective and the layer-specific nature of these training methods precludes the possibility of supervised learning. Therefore, semi-supervised learning, which leverages both labeled and unlabeled data through alternating unsupervised and supervised learning phases, cannot be easily performed \citep{van2020survey, zhu2022introduction}.

In this study, we introduce a simple method, illustrated in Figure \ref{fig:unsupervised-end2end}(C), that allows a network designed for end-to-end supervised learning to also efficiently achieve unsupervised learning. Leveraging the network's inherent dynamics, whether recurrent or feedforward, in response to input data we generate a self-defined target for each data point using a Winner-Take-All mechanism in the last layer, derived from biological principles \citep{maass2000computational}. Additionally, we apply homeostasis, another bio-inspired approach, to regulate the labeling process and prevent feature collapse, where the model outputs identical predictions for all inputs \citep{Querlioz2013}. 
These bio-inspired approaches, optimized through evolution, are inherently resource-efficient and therefore particularly valuable for hardware design.
On one hand, compared to self-supervised learning, our unsupervised target depends only on the current network output and the homeostasis term, requiring minimal memory and thus being more compatible with edge AI hardware. On the other hand, we use a global objective that accommodates both global (Backpropagation) and local (Equilibrium Propagation) learning rules, retaining the advantages of end-to-end supervised learning while also enabling online unsupervised and semi-supervised training of neural networks.

\begin{figure}[ht]
    \centering
    \includegraphics[width=0.6\textwidth]{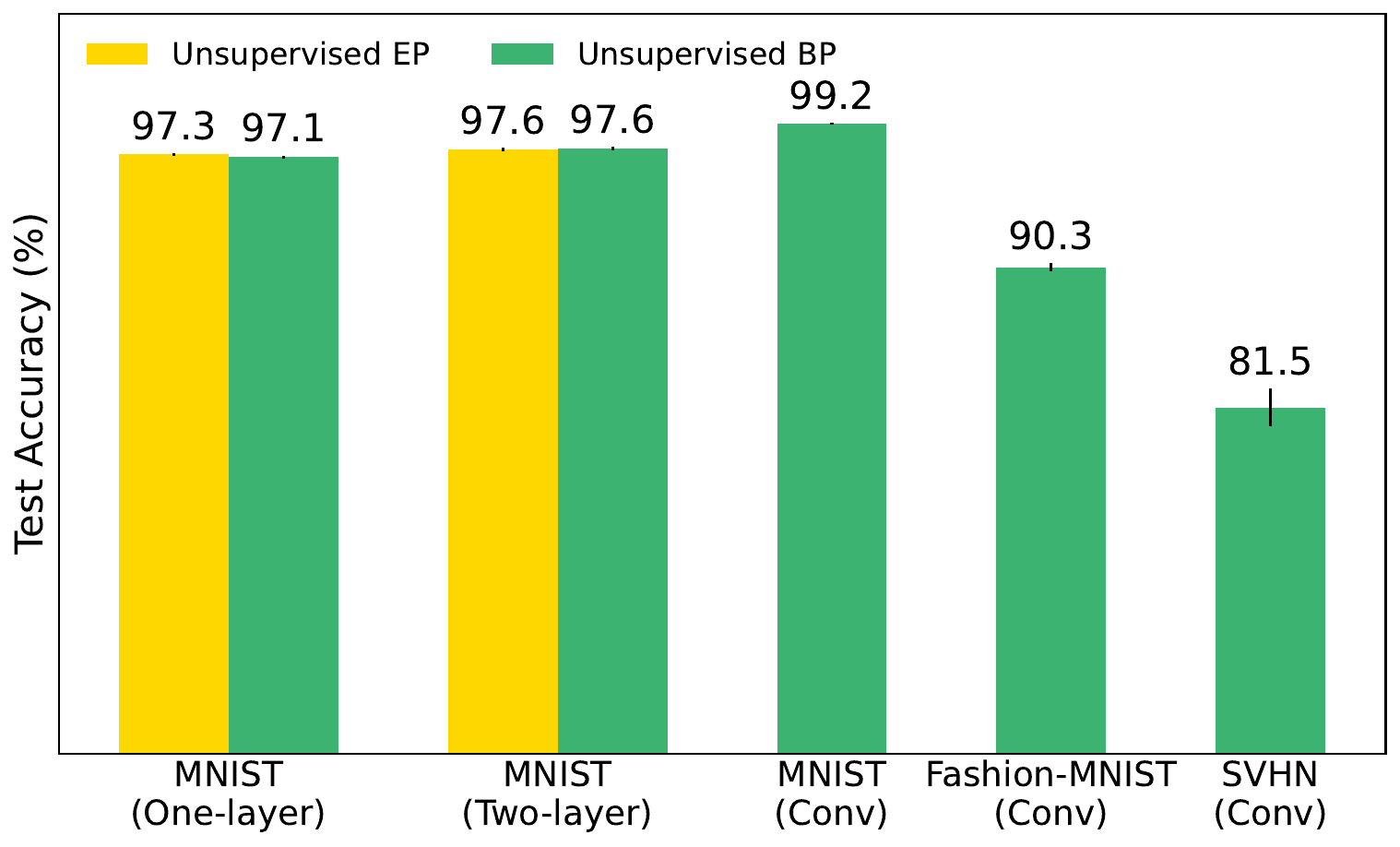}
    \footnotesize
    \caption{\label{fig:main_results} Test accuracies achieved using our Unsupervised End-to-End Training Method for weight updates, utilizing Backpropagation and Equilibrium Propagation, applied to fully-connected and convolutional networks across the MNIST, Fashion-MNIST, and SVHN datasets.}
\end{figure}

With our versatile and framework-agnostic we demonstrate the first results of unsupervised and semi-supervised training with Equilibrium Propagation \citep{scellier2017equilibrium, ernoult2019updates, ernoult2020Ceqprop, laborieux2021scaling,  laborieux2022holomorphic, scellier2024energy}. This algorithm, originally applied to supervised learning, shows great promise for edge AI implementations on analog hardware because it computes weight updates solely through the dynamics of the hardware and with local information only - paving the way for implementations requiring less memory and circuitry \citep{Yi2022, dillavou2022demonstration, kendall2020training, laydevant2023training, martin2021eqspike}. We also demonstrate the method's effectiveness in multilayer and convolutional networks networks trained with conventional non-local backpropagation. The results for the different datasets are summarized in Figure \ref{fig:main_results}.

More specifically, the contributions of this work are the following:
\begin{itemize}

  \item Our innovation is a special objective function that takes inspiration from both SOTA Self-Supervised Learning methods and biological mechanisms. 
  Information about the input data in the representation vector (output of the model) is sparsified with a Winner-Take-All mechanism while the features are regularized through a homeostasis mechanism, which imposes the neurons to have a certain level of variance given different input data.

  \item We first perform unsupervised learning on the MNIST dataset with a multilayer perceptron, within two different frameworks: a bidirectional recurrent network with static inputs trained by Equilibrium-Propagation (EP) and a feed-forward network trained by Backpropagation (BP). Using a fully-connected network with a hidden layer, we achieve a test accuracy of 97.6\% with both EP and BP (see Figure \ref{fig:main_results}. The analysis of the classification accuracy and the quality of the learned features reveal the improvement in performance of the network with a hidden layer compared to a single-layer network.

  \item We show that our Self-Defined Unsupervised Target method trains Convolutional Networks in an unsupervised way through BP with competitive test accuracies on MNIST  (99.2\%), Fashion-MNIST (90.3\%) and the RGB color dataset SVHN (81.5\%) (Figure \ref{fig:main_results}).
  
  \item We demonstrate that we can alternate unsupervised and supervised learning phases on the same network, by adapting the target type based on the availability of labels. Our two-layer network thus achieves a test accuracy of 96.6\% training with only 600 labeled samples from the MNIST dataset, outperforming semi-supervised methods based on 'pseudo-labels' \citep{Lee2013}.

\end{itemize}


\section{Methods}


\subsection{Self-defined Unsupervised Target}

We introduce a self-defined target $d$ to determine an unsupervised global loss from the network output, by a direct modification of the supervised global loss.
For this, we equip the output layer of the neural network with two bio-mechanisms, Winner-Take-All (WTA) selectivity and homeostasis regularization, where WTA encourages output neurons to respond differently to each input, and homeostasis ensures that the average unsupervised target value across the examples is balanced among the output neurons \citep{Querlioz2013, bichler2012extraction, miconi2021hebbian}.

For each new unlabeled example presented at the input of the neural network, we select the 'winners' as the top $k$ neurons in the output layer $y$ with the highest response above a threshold defined by the homeostasis penalty term $H$. For each output neuron $i$, the target value of the 'winners' $d_i$ is set to `1', while the target value of the other neurons is set to `0':

\begin{equation}
    d_i = \left\lbrace
    \eqalign{
       1 \quad & \mbox{if} \; y_i - H_i\in \{ \mbox{$k$ largest  elements of $y - H$}\},\cr
        0 \quad &\mbox{otherwise}.
       }
     \right.
    \label{equa:unsupervised_target}
\end{equation}
The vectors $y$ and $H$ have the same length $n_{output}$, the number of output neurons. The term $H$ may take on positive or negative values to either encourage or suppress neuronal competition, respectively. The homeostasis value is set to zero at the beginning. In the pseudo-code Algorithm.~\ref{pseudocode}, Eq.~\ref{equa:unsupervised_target} is shortened as $d = \mbox{kWTA}(y-H)$.

The WTA competitive environment enables neurons to specialize in specific data patterns. When data from different classes activate distinct output neurons, data clustering is obtained. However, without a regulatory mechanism for neuron activity during training, dominant neurons may respond to all inputs, while the others become silent. The homeostasis penalty term $H$, introduced in Eq.~\ref{equa:unsupervised_target}, addresses the issue of silent neurons by ensuring that all output neurons achieve an equal 'winning frequency.' Consequently, each neuron learns from, on average, the same number of images presented at the input. Following the proposal of \cite{Querlioz2013}, the homeostasis term $H$ is updated as 
\begin{equation}
    \dot{H} \leftarrow \gamma (\langle d \rangle - T),
    \label{equa:homeostasis}
\end{equation}
where $\langle d \rangle$ is the average of the unsupervised target across examples, $T$ the desired winning frequency defined as the ratio between the number of winner neurons $k$ and the number of output neurons $n_{output}$, and $\gamma$ a hyperparameter. The average activity, $\langle d \rangle$, is defined differently depending on whether the input data is batched or not. In sequential mode, where inputs are sent to the network one by one, it is defined as the exponential moving average of the unsupervised target. In mini-batch mode, where inputs are processed in batches, it is defined as the average unsupervised target across the batch. The equations for these definitions (Eq. \ref{equa:EMA} and Eq. \ref{equa:CMA}) and additional details can be found in the Appendices.

Once the unsupervised target is defined, we calculate the unsupervised loss $L$ as the Mean Squared Error (MSE) between the output $y$ and unsupervised target $d$: 
\begin{equation}
    L = \frac{1}{N} \sum_{n\in \textbf{N}}(\textbf{y}_n - \textbf{d}_n)^2
    \label{equa:unsupervised_loss},
\end{equation}
where $\boldsymbol{N}$ designates the dataset with $N$ samples, and $\boldsymbol{y}_n$ and $\boldsymbol{d}_n$ are respectively the output value and the unsupervised target corresponding to the input $n$.
For the training process, the weights of the network are updated by performing stochastic gradient descent on $L$. In the case of BP, the gradients are calculated directly by using the chain rule; in the case of EP, they are approximated by local neural activities. 

The complete unsupervised training process is described in Algorithm \ref{pseudocode}, where $X_{train}$ is the training dataset, $\textit{Model}$ the neural network model, $W$ the training parameters, $lr$ the learning rate, and $N$ the epoch number.

\begin{algorithm}
\footnotesize
\caption[Pseudo_code]{Pseudo Code for Unsupervised End-to-End Training with Self-Defined Target}
\begin{algorithmic}[1]
     \State \textbf{Inputs:} 
    \State \hspace{\algorithmicindent} $X_{train}$, \;$\textit{Model}$, \;$W$, \;$\gamma$, \;$lr$, \;$N$, \; $L$, \; T. 
\State
    
    \For{epoch $n \in N$}
        \State Initialize the homeostasis term $H \leftarrow 0$
        \For{each image $x \in X_{train}$}
            \State Get output : $y \leftarrow \textit{Model}(x)$
            \State Define unsupervised target : $d \leftarrow \mbox{kWTA}(y - H)$ \Comment{\textbf{Target decision}}
            \State Calculate loss: $l \leftarrow L(y,d)$
            \State Calculate gradient: $\Delta W \leftarrow - \frac{\partial l}{\partial W}$ \Comment{Analytically (BP) or via local activities (EP)}
            \State Update weight: $W \leftarrow W + lr\cdot\Delta W $ 
            \State Update the homeostasis term $H \leftarrow H + \gamma (\langle d \rangle - T)$
        \EndFor
    \EndFor
\end{algorithmic}
    \label{pseudocode}
\end{algorithm}

In the pseudo-code and experiments presented here, the homeostasis term $H$ is initialized at each epoch, but we found that the training performance on MNIST is not affected if it is not the case.

\setlength{\belowcaptionskip}{5pt}
\begin{table}[hb]
\caption{\label{tab:architecture_notation}Notations used in our neural network models.}
\begin{minipage}{\textwidth}
    \centering
    \resizebox{1\columnwidth}{!}{
    \footnotesize
    \begin{tabular}{p{0.15\textwidth}|p{0.22\textwidth}|p{0.6\textwidth}}
    \toprule
    Training type & Notation & Definition\\
    \midrule
    &One-layer network & Unsupervised neural network consisting of a single synaptic layer, directly linking the input to the output without any hidden layer.\\
    \cmidrule{2-3}
   &Two-layer network & Unsupervised neural network consisting of two processing layers, linking the input to the output with one hidden layer.\\
    \cmidrule{2-3}
   Unsupervised Training&Unsupervised output layer & Last layer of the unsupervised neural network, where the unsupervised loss is defined.  \\
   \cmidrule{2-3}
   &Unsupervised hidden layer & Middle layer connecting the input and output layer in the unsupervised two-layer network.\\
   \cmidrule{2-3}
   &Readout layer & Linear classifier layer added to the top of the unsupervised network, integrated subsequent to the completion of unsupervised training.\\
   \midrule
   &Multi-layer perceptron (MLP) & Fully connected neural network used for semi-supervised training, consisting of at least one hidden layer.\\
   \cmidrule{2-3}
Semi-supervised training & Output layer & Last layer of the MLP, where the unsupervised or supervised loss is defined. Neuron count corresponds to the number of classes in the dataset.\\
   \cmidrule{2-3}
   & Hidden layer & The middle layer between the input layer and output layer in the MLP.\\
   \bottomrule
\end{tabular}
}
\end{minipage}
\end{table}


\subsection{Training Methodology}

A comprehensive list of all layer and network structure notations is presented in Table~\ref{tab:architecture_notation}.


\subsubsection{Unsupervised Learning}

We train two distinct network architectures using our unsupervised learning framework: a one-layer network and a two-layer network. These notations discount the input layer, as its values are clamped to the inputs. The unsupervised loss is always computed at the final layer, designated as the unsupervised output layer. In the two-layer network, the first layer is called the unsupervised hidden layer.

Class association is used exclusively for unsupervised training to evaluate classification performance. This procedure is implemented at the end of the unsupervised training phase, prior to the testing phase. 
During the association phase, the weights of the trained unsupervised network are kept frozen. In our approach, we utilize two distinct methods: {direct association} and a {linear classifier}. Both methods require a subset of labeled data. However, at no point are the weights of the unsupervised network updated using these labels.

In the {direct association} method, each output neuron of the neural network is assigned to the class to which it shows the strongest response. This technique is widely employed in unsupervised training \citep{Querlioz2013, DiehlCook2015, moraitis2022softhebb}. We expose the network to a subset of labeled data and record the network response to each input. For every output neuron, we determine its average response to the different classes. The class that elicits the highest average response from a neuron is designated as that neuron's represented class. During the testing phase, the class whose corresponding neurons exhibit the highest average response is then identified as the predicted class for the given input.

In the {linear classifier} method, a linear classifier is trained in a supervised manner on a subset of labeled data \citep{Ferre2018WTA, miconi2021hebbian, moraitis2022softhebb}. It is built on top of the unsupervised network, using the unsupervised network's output as its input. Upon completion of training, each class is represented by a distinct output neuron in the classifier. The prediction of the class is determined by the output neuron that exhibits the maximum value. In our notation, the linear classifier added for classification is not included in the total layer count of the network. 

Figure~\ref{fig:unsupervised_structure_process} illustrates the unsupervised network's architecture and the training-testing workflow, using a two-layer network as an example.


\subsubsection{Semi-supervised Learning}

For the semi-supervised learning framework, we employ a {multilayer perceptron (MLP)} with a single hidden layer, trained using a limited amount of labeled data. The MLP consists of two fully connected layers: a \textbf{hidden layer} and an \textbf{output layer}. Both the supervised and unsupervised losses are computed at the output layer.

Unlike fully unsupervised learning, our semi-supervised framework does not involve a class association step. Training initially proceeds in a supervised manner, followed by unsupervised learning. Therefore, the output layer's neuron count corresponds to the number of classes in the dataset. Upon completion of the training phase, the network is directly subjected to the testing process. The structure of the semi-supervised network, along with its configuration, is shown in Figure \ref{fig:semi_supervised_structure_process}.


\subsection{Additional Regularization Methods}

For unsupervised training, dropout \citep{Dropout_initial} is applied at both the input and output layers of our neural network architecture, with dropout probabilities set to 0.3 and 0.2. Applying dropout to the output layer helps to introduce randomness in the unsupervised target decision. This contrasts with supervised learning, where dropout is usually not used at the output layer, as the number of output neurons typically matches the number of classes. For our semi-supervised training, where the number of output neurons also matches the number of classes, dropout is not employed at the output, but at the input and hidden layers, with probabilities of 0.3 and 0.5 respectively to prevent over-fitting.

In the case of EP training, we also use label smoothing \citep{labelsmooth2017} to penalize overconfidence in the unsupervised targets. Instead of pushing output neurons to the values of `1' and `0', the $k$ winner neurons are pushed to a less confident value, while the other neurons receive a less strong inhibition value. 

These two additional regularization methods are detailed in the Appendices.


\section{Results}


\subsection{Unsupervised Learning}

\begin{figure}[ht]
    \includegraphics[width=1\textwidth]{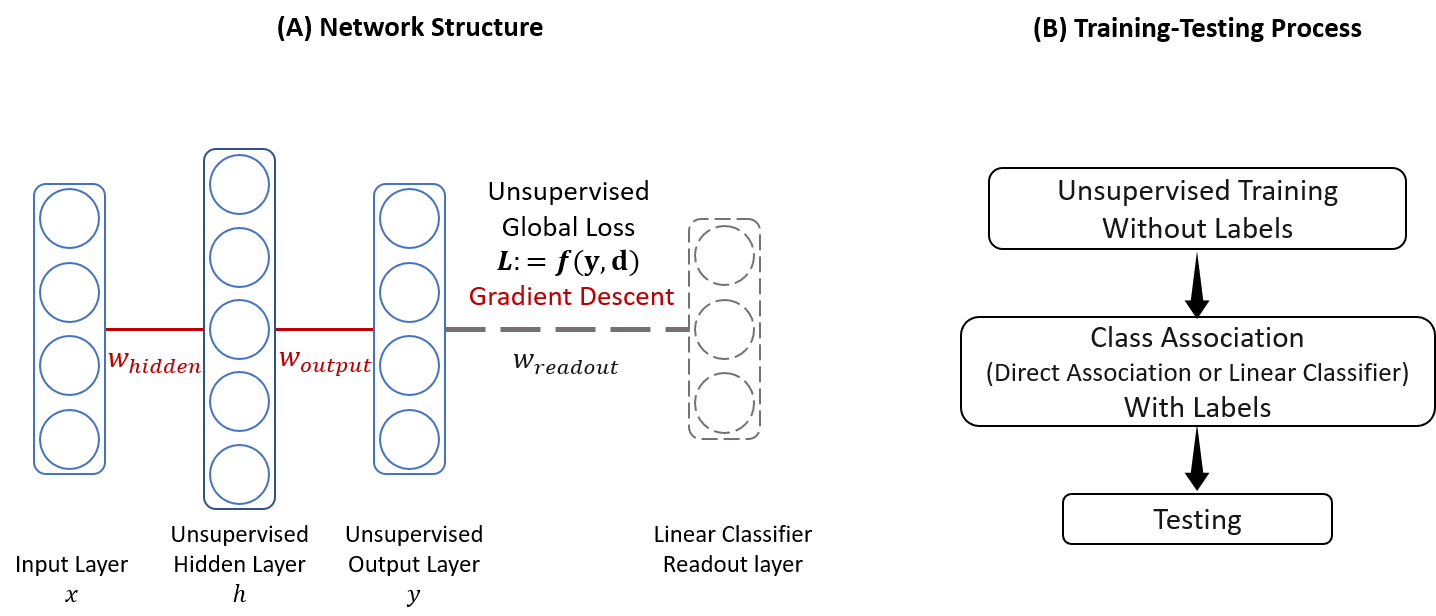}
    \caption{\label{fig:unsupervised_structure_process} Unsupervised learning: \textbf{(A)} Network architecture and \textbf{(B)} Training-Testing process}
\end{figure}

In this section, we present simulation results for the MNIST handwritten digits dataset \citep{lecun2010mnist}, utilizing both Backpropagation (BP) and Equilibrium Propagation (EP) to train the networks with our unsupervised global loss. The general network structure is shown in Figure \ref{fig:unsupervised_structure_process}(A). In practice, BP is adapted to train a standard feed-forward network, while training with EP requires a bidirectional network. In this bidirectional recurrent network, hidden neurons receive information from both the input neurons and the output neurons, contrary to the feed-forward network where hidden neurons receive information only from the input neurons. In this convergent recurrent structure, neural values evolve towards stable values for a duration corresponding to their relaxation time after each given input is applied.

We implement both unsupervised EP and BP on two distinct networks: a one-layer network and a two-layer network, following the notations in Table \ref{tab:architecture_notation}, and compare their performance. For the one-layer network, since there are no hidden neurons, there is no difference between the feed-forward and EP-used bidirectional networks.
All the reported results represent the averaged values from ten simulations.

 The sequence of training, class association, and testing steps, described in the Methods section, is illustrated in Figure \ref{fig:unsupervised_structure_process}(B).

\begin{figure}[htbp]
    \centering
    \includegraphics[width=1\textwidth]{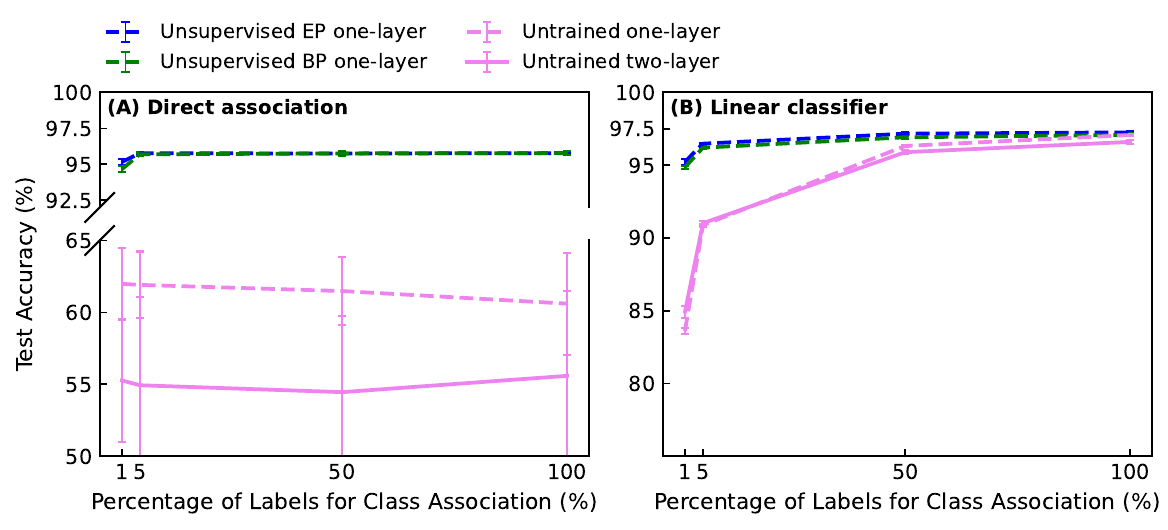}
    \caption{\label{fig:onelayer_compare_untrained}
    Test accuracy on MNIST for unsupervised learning, as a function of the percentage of labeled data used for class association: \textbf{(A)} with direct association; \textbf{(B)} with a linear classifier. Green dotted line: one-layer network trained by unsupervised BP, blue dotted line: one-layer network trained by unsupervised EP. Pink lines: untrained networks, including one-layer network (dotted line) and two-layer network (solid line). The single-layer network has 2,000 output neurons, while the two-layer version adds 2,000 hidden neurons.  
    }
\end{figure}

Figure \ref{fig:onelayer_compare_untrained} shows the test accuracy on MNIST obtained by training a one-layer neural network with our unsupervised method through EP and BP. The results are compared to untrained (i.e., random weights) one-layer and two-layer networks. The results are plotted as a function of the percentage of labeled data used for class association in the MNIST database, for direct association (A) and with a linear classifier (B). For both association methods, the unsupervisedly trained networks exhibit a marked increase in accuracy compared to the untrained networks. The accuracy obtained with EP and BP is comparable. 

The results obtained with the two class association methods are different. For direct association (Figure \ref{fig:onelayer_compare_untrained}(A)), less than 5\% of labels are needed to reach the peak accuracy of about 95.8\%, beyond which the use of more labeled data does not significantly impact the accuracy. On the contrary, as shown in Figure \ref{fig:onelayer_compare_untrained}(B), the performance of the linear classifier improves significantly with increasing amounts of labeled data, underscoring the need for substantial labeled data to effectively train its extensive parameters in a supervised manner. The highest accuracy for the one-layer network, close to 97.2\% is obtained when all labels in the database are used. 

The performance of the untrained networks in Figure \ref{fig:onelayer_compare_untrained}(B) also strongly increases with the amount of labeled data used to train the linear classifier, reaching the accuracy of the trained networks when all the labels are leveraged. With direct association in Figure \ref{fig:onelayer_compare_untrained}(A) on the other hand, the gap between the trained and untrained networks remains large and constant versus the amount of labeled data. Therefore, while the accuracy results are reduced with the direct association method, it provides a better framework than the linear classifier to evaluate the contribution of unsupervised training to the network accuracy, compared to untrained networks.

\begin{figure}[htbp]
    \centering
    \includegraphics[width=1\textwidth]{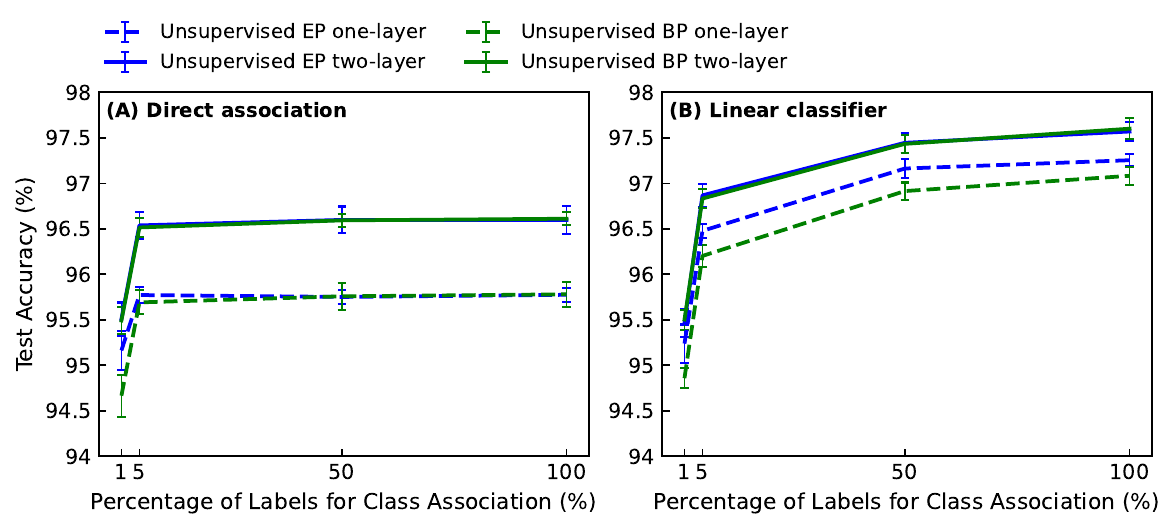}
    \caption{
    Test accuracy on MNIST for unsupervised training as a function of the amount of labeled data used for class association: \textbf{(A)} with direct association; \textbf{(B)} with a linear classifier. The two-layer network trained by unsupervised BP (green solid line) and EP (blue solid line) is compared with the one-layer unsupervised trained network (dotted lines). The single-layer network has 2,000 output neurons, while the two-layer version adds 2,000 hidden neurons.
    }
    \label{fig:twolayer_improve}
\end{figure}

In Figure \ref{fig:onelayer_compare_untrained}(A) and (B), we see that the two-layer untrained network has a decreased accuracy compared to the one-layer untrained network for both class association methods. We now evaluate in Figure \ref{fig:twolayer_improve} the impact of training the two-layer neural network with our method compared to a one-layer network. As previously, the accuracy is plotted against the amount of labeled data used in the class association process, for direct association in \ref{fig:twolayer_improve}(A) and with the additional classifier in \ref{fig:twolayer_improve}(B). In all cases, the trained two-layer networks achieve a higher accuracy than the trained one-layer networks: The addition of a hidden layer consistently improves the classification ability. The enhancement is stronger in the direct association method than with the linear classifier: The optimization of the classifier's parameters reduces the impact of unsupervised training on the overall accuracy.

The highest accuracy, of 97.6\%, is achieved by the two-layer network trained with EP or BP, using a linear classifier and all available labeled data for association (solid lines in Figure \ref{fig:twolayer_improve}(B)). 
Overall, the enhancement in accuracy obtained by adding a hidden layer in the networks indicates the effectiveness of the proposed end-to-end unsupervised training: In traditional unsupervised Hebbian or STDP learning methods, adding a hidden layer in a fully connected network typically does not lead to improvements in accuracy \citep{zhou2022activation, journe2022hebbian}.

\begin{figure}[htbp]
    \centering
    \includegraphics[width=0.95\textwidth]{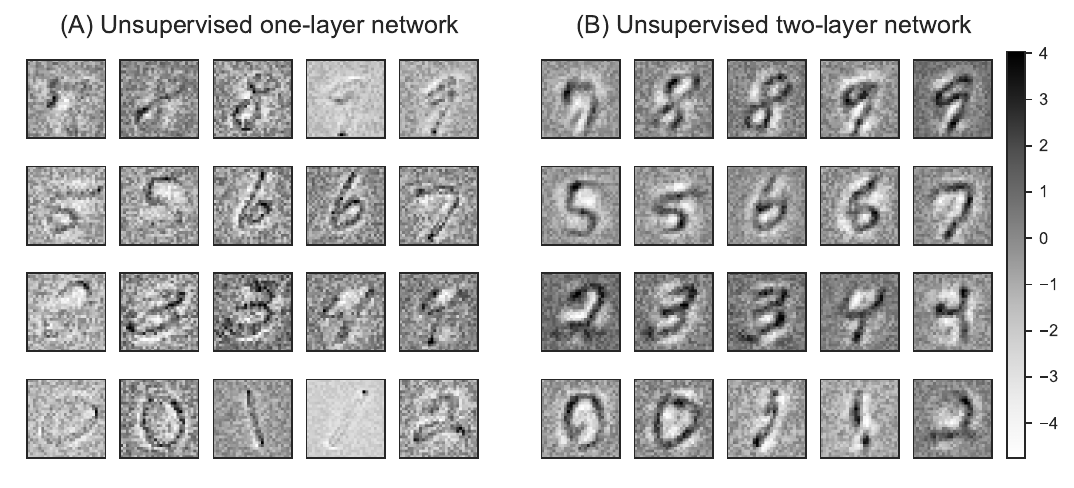}
    \caption{
    Maximum activation images: input images that elicit the strongest responses in 20 selected (the two highest responses for each class) output neurons of the networks trained by unsupervised BP on the MNIST dataset. \textbf{(A)} One-layer network \textbf{(B)} Two-layer network. Each network comprises 2,000 output neurons. The hidden layer of the two-layer network consists of 2,000 neurons. 
     }
    \label{fig:visu_compare_BP}
\end{figure}

We use a visualization method to further study the impact of the hidden layer in our fully-connected network trained in an unsupervised way \citep{Erhan2009VisualizingHF}. This technique finds through optimization the input images that maximize neuronal responses in the trained network, producing 'maximum activation images'. These images provide insights into the learned features of each neuron in the network. Figure \ref{fig:visu_compare_BP} showcases the maximum activation images for output neurons from two networks trained via unsupervised BP: with a single unsupervised layer (Fig.\ref{fig:visu_compare_BP} (A)) and with two unsupervised layers (Fig.\ref{fig:visu_compare_BP} (B)). Compared to the single-layer network, the two-layer network displays digit features with clearer outlines and sharper contrast against the background. The single-layer approach tends to focus on the extreme values of digit pixels, occasionally omitting parts of the digit features. In contrast, the two-layer unsupervised network captures the entirety of the digit structure. This observation provides an explanation for the improved accuracy obtained by incorporating the hidden layer. The inclusion of a hidden layer results in clearer feature representation at the output layer and enhanced accuracy. 

\begin{figure}[htbp]
    \centering
    \includegraphics[width=0.8\textwidth]{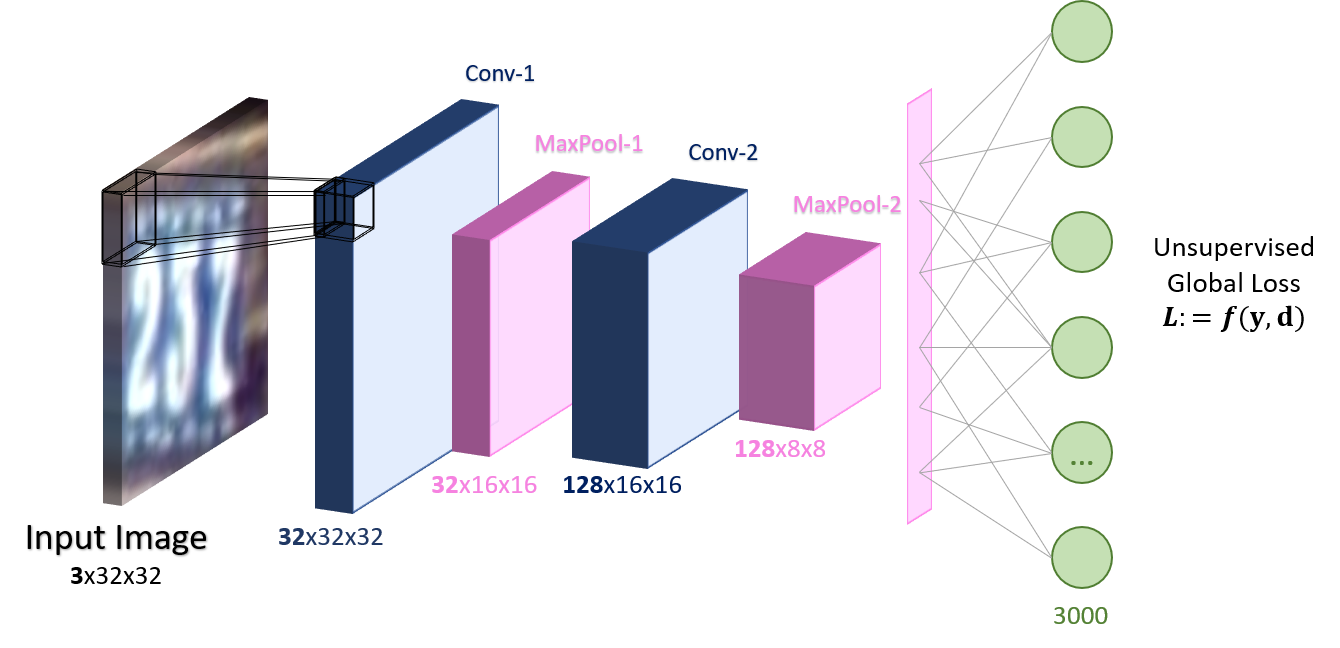}
    \caption{Architecture of the convolutional network, consisting of two convolutional layers followed by a flatten operation and a fully connected layer.}
    \label{fig:cnn_results}
\end{figure}

We further applied our algorithm to convolutional neural networks (CNNs) for the MNIST dataset, as well as two more complex datasets: Fashion MNIST \citep{xiao2017fashion} and Street View House Numbers (SVHN) \citep{Netzer2011svhn}. For SVHN training, we did not use the extra training set. As shown in Figure \ref{fig:cnn_results}, our convolutional network is composed of two convolutional layers followed by a fully connected layer. The input data is passed through the convolutional layers, then flattened and sent to the fully connected layer. After each convolutional layer, max pooling is applied to downscale the image before passing it to the next layer. The unsupervised loss is calculated at the output of the fully connected layer. All details about the CNN architecture can be found in the appendices. After unsupervised training, we trained a linear classifier with full labels to evaluate the test accuracy.

For the MNIST dataset, our CNN achieved a test accuracy of 99.2\%, which is 1.6\% higher than the result obtained with two unsupervised layers. This demonstrates that our method can be integrated into deeper networks and benefit from the convolutional architecture. Our method also generalizes well to more complex datasets, achieving 90.3\% accuracy on Fashion-MNIST and 81.5\% accuracy on the SVHN dataset.
These results compete with other fully unsupervised training methods across all datasets, as detailed in Section \ref{sec:cnn_compare} and in Table \ref{tab:cnn_compare}.
They also approach those of purely supervised backpropagation training results for the MNIST (99.2 \% for our unsupervised BP vs. 99.6\% for supervised BP)and Fashion-MNIST datasets (90.3 \% for our unsupervised BP vs. 93.1\% for supervised BP). For the SVHN dataset, supervised backpropagation achieves a test accuracy of 93.4\% using the same CNN architecture, indicating that further improvements are needed to enhance our method for more complex tasks.

\begin{table}[H]
    \caption{\label{tab:dataset} Test accuracy of our convolutional network trained with unsupervised backpropagation across three different datasets: MNIST, Fashion-MNIST, and SVHN}
    \begin{minipage}{\textwidth}
    \centering
    \resizebox{.9\columnwidth}{!}{
    \begin{tabular}{ccccccc}
    \toprule
 Dataset & Data type & Input size & Training samples & Test samples & Our Test Accuracy\\
    \midrule
    MNIST & Grey scale & 28x28 & 60,000 &  10,000  &  99.17($\pm$ 0.05)\%\\
    Fashion-MNIST & Grey scale & 28x28  & 60,000  & 10,000   & 90.25($\pm$ 0.26)\%\\
    SVHN &  RGB & 3x32x32  &  73,257 & 26,032  & 81.53($\pm$ 1.15)\%\\
    \toprule
    \end{tabular}
    }
\end{minipage}
\end{table}

\subsection{Semi-supervised Learning}

In this section, we address a scenario where only a limited amount of training data is labeled, and the majority remains unlabeled, which is the case in practical situations. Semi-supervised learning serves as a pertinent strategy under these circumstances \citep{thomas2009semi, van2020survey}. On one hand, the inclusion of some labeled data can substantially enhance accuracy for unsupervised clustering tasks. On the other hand, the vast volume of unlabeled data, when aligned with the predictions of a supervised network, aids in generalizing the training outcomes. 

In our approach, as our unsupervised global loss retains the same form as the supervised loss, we can leverage the benefits of both labeled and unlabeled data by adapting the target decision. For analog devices suitable for supervised Equilibrium Propagation (EP) training \citep{Yi2022, dillavou2022demonstration, kendall2020training, laydevant2023training}, our method enables semi-supervised training by modifying the target decision without any hardware changes.

\begin{figure}[ht]
    \centering
    \includegraphics[width=1\textwidth]{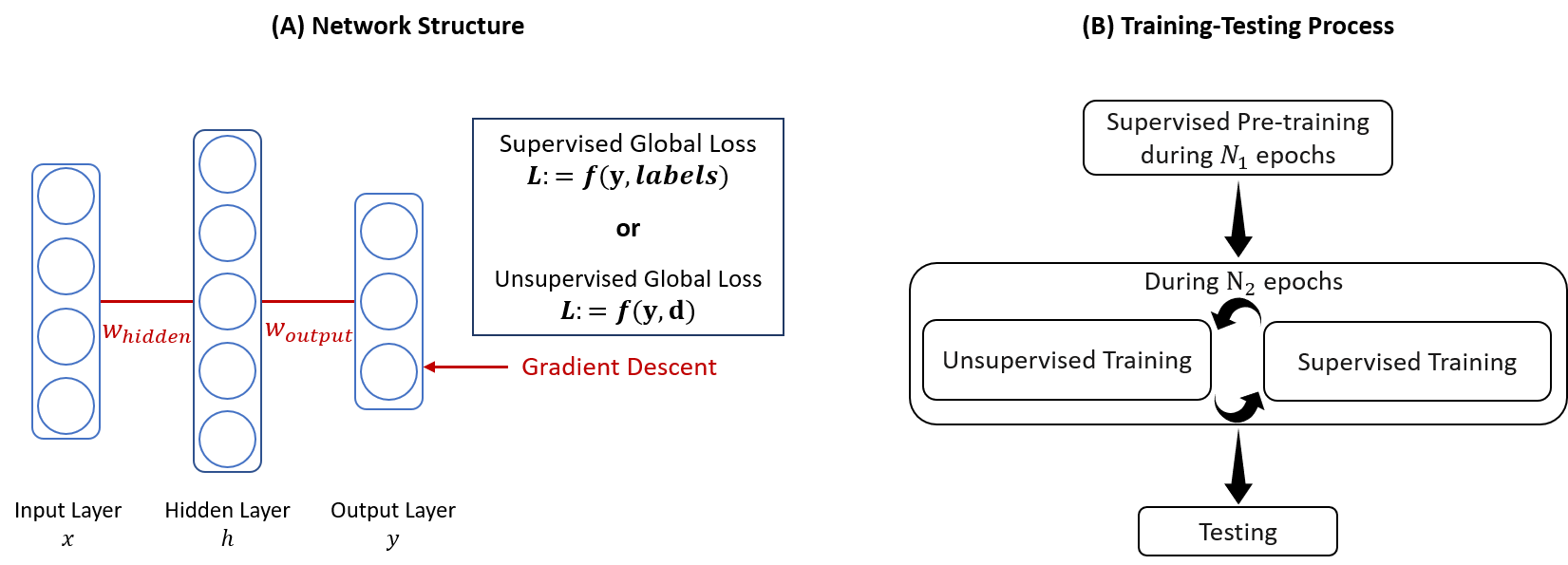}
    \caption{Semi-supervised Learning: \textbf{(A)} Network architecture and \textbf{(B)} Training-Testing process.}
\label{fig:semi_supervised_structure_process}
\end{figure}

Figure \ref{fig:semi_supervised_structure_process} outlines our semi-supervised learning methodology, introduced in Methods, which aligns with standard practices in machine learning \citep{yarowsky1995unsupervised, zhu2022introduction}. Our method starts with the model undergoing training using a select, small subset of labeled data in a phase known as supervised pre-training, for $N_1$ epochs. The duration of $N_1$ is set to a magnitude where extending the number of epochs beyond this point yields no further enhancement in the network's efficiency. Starting with supervised, rather than unsupervised learning allows matching the number of output neurons to the number of classes, thus significantly reducing their number, together with requirements on memory and computational resources. After the initial supervised training, we transition to unsupervised learning with the self-defined target. Following each unsupervised training epoch, we perform one epoch of supervised training with the same set of labeled data as used initially. This supervised re-training ensures that the network retains its earlier supervised learning insights. This iterative process persists for $N_2$ epochs. The hyperparameters $N_1$ and $N_2$ are specified in the Appendices.

In semi-supervised learning, the use of a homeostasis term is, in principle, not compulsory in the self-defined target expression of Eq:~ \ref{equa:unsupervised_target}: The supervised pre-training ensures that each output neuron becomes specialized in one class, averting the issue of silent neurons. In the following, we compare the results obtained with and without homeostasis.

Given that the number of output neurons corresponds to the number of classes in both methods, the number of winner neurons $k$ is consistently set to 1.

The network architecture used in these experiments is a multi-layer perceptron with 5,000 hidden neurons and 10 output neurons, each corresponding to a class prediction. Experiments were conducted using both BP and EP algorithms, and to mitigate overfitting, dropout was incorporated during both training phases.

\begin{figure}[htbp]
    \centering
    \includegraphics[width=1\textwidth]{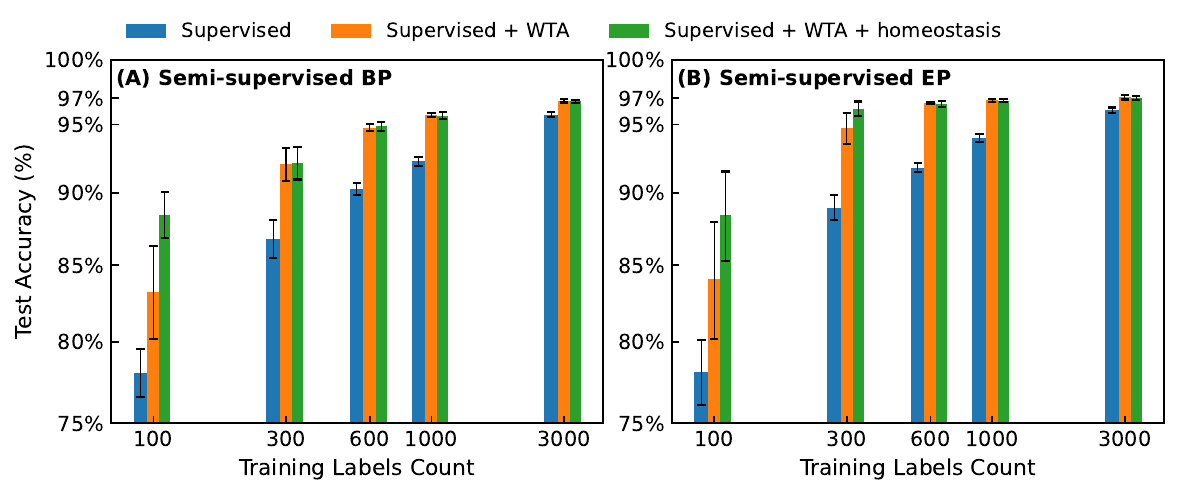}
    \caption{Test accuracy (\%) of semi-supervised methods on the MNIST dataset using \textbf{(A)} BP and \textbf{(B)} EP for loss optimization. The network is firstly trained by limited labeled data, then trained in a unsupervised manner by two different unsupervised targets, shown as '+ WTA' and '+ WTA + Homeostasis' respectively. }
    \label{fig:Semi_BPEP}
\end{figure}

Figure \ref{fig:Semi_BPEP} shows the test accuracy on the MNIST dataset as a function of the quantity of labeled training data, for BP (Figure \ref{fig:Semi_BPEP}(A) and EP (Figure \ref{fig:Semi_BPEP}(B)). Our analysis encompasses five scenarios with different amounts of available labeled data: 100, 300, 600, 1000, and 3000 labels. This labeled data is randomly selected from the training dataset, ensuring an equal representation of each class. To account for variability in labeled data selection, each scenario is subjected to multiple simulation runs: thirty for the 100-label scenario due to high variance in its results, fifteen for the simulation with 300 labeled images, and ten for the others. 

Figure \ref{fig:Semi_BPEP} compares the accuracy of semi-supervised learning without and with homeostasis against purely supervised learning. The addition of unsupervised training markedly enhances performance, mainly because supervised training struggles to generalize from limited labeled data, often leading to overfitting. Unsupervised training mitigates this by utilizing the wealth of unlabeled data. When fewer than 600 labels are available, the homeostasis regularization becomes crucial. This is logical, as limited labeled data renders the WTA-defined target, essentially the prediction accuracy from supervised training, less accurate. Thus, imposing a winning frequency balance helps sharpen these predictions. Additionally, EP surpasses BP in semi-supervised settings, likely due to its quicker convergence and the augmented impact of label smoothing in recurrent networks.

\begin{figure}[htbp]
    \centering
    \includegraphics[width=1\textwidth]{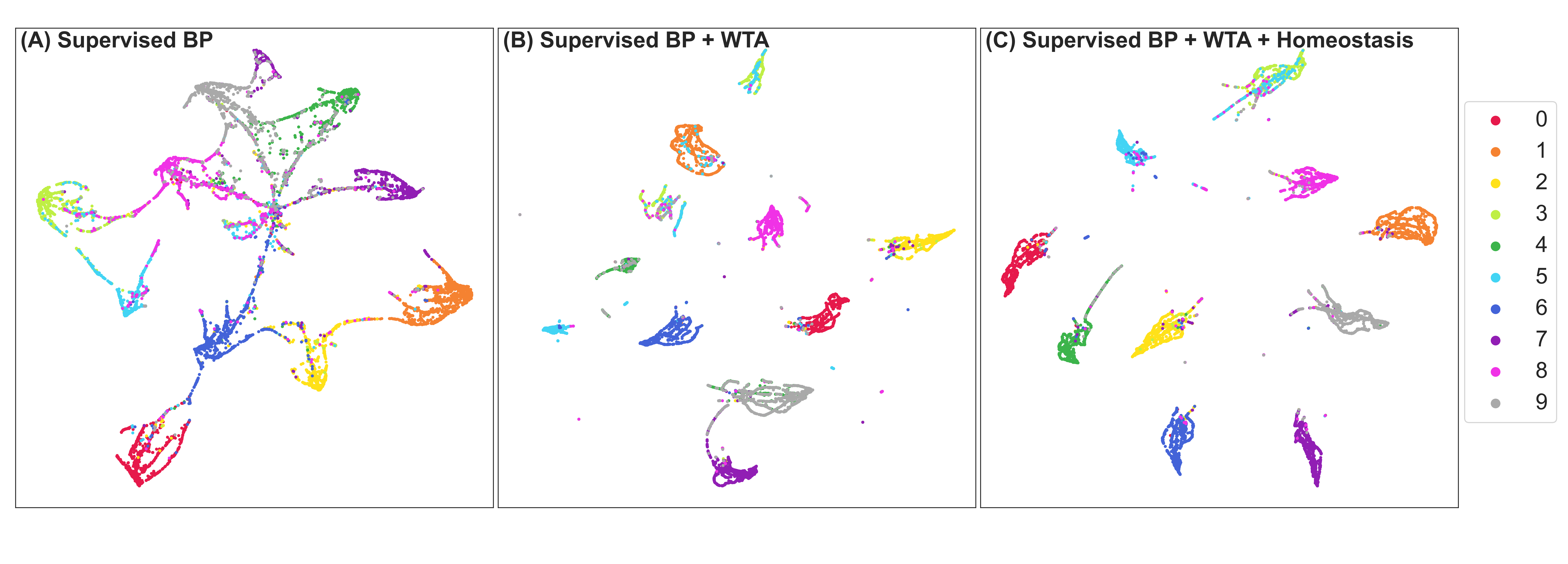}
    \vspace{-20pt}
    \caption{UMAP 2D embedding of the semi-supervised network output on MNIST test data \citep{mcinnes2018umap}. \textbf{(A)}: Output of the initial supervised training on 100 labeled images; \textbf{(B)}: Output of unsupervised training with WTA selectivity after the initial supervised step; \textbf{(C)}: Output  unsupervised training with both WTA and homeostasis after the initial supervised step. 
    }
    \label{fig:UMAP_semi_BP}
\end{figure}

We use the Uniform manifold approximation and projection for dimension reduction (UMAP) method \citep{mcinnes2018umap} to visually assess the influence of WTA selectivity and homeostasis by condensing the network output for MNIST test data into a two-dimensional space. The corresponding 2D embedded output after initial supervised training on 100 labeled samples is showcased in Figure \ref{fig:UMAP_semi_BP}(A). Subsequent outputs, after unsupervised training with WTA and the combination of WTA with homeostasis, are depicted in sub-figures (B) and (C), respectively. As we introduce WTA and its combination with homeostasis, the network output grows increasingly compact and more distinguishable: This visualization underscores that incorporating unlabeled data enhances class differentiation based on prior supervised training outcomes. Moreover, homeostasis demonstrates its benefits even within semi-supervised training scenarios.

Overall, this study demonstrates that unsupervised learning realized through our self-defined target can integrate effectively with supervised learning to enhance model generalization from abundant unlabeled data. This semi-supervised approach is not only efficient but straightforward to implement. Compared to purely unsupervised learning, the incorporation of even a small amount of labeled data in the training significantly reduces the number of output neurons and eliminates the need for a class association process, resulting in a more hardware-friendly solution for edge training.


\section{Discussion}


\subsection{Single-layer Unsupervised Network: Comparison with Hebbian Learning}

The scenario of a single-layer unsupervised network provides a straightforward comparison between our proposed weight updating rule and the conventional Hebbian learning rule. Consider the synaptic weight $W_{ij}$ connecting neuron $i$ in the input layer to neuron $j$ in the hidden layer. Let $x_i$ denote the input value for neuron $i$, and 
$y_j$ the aggregated input for neuron $j$ before the activation $\rho$, i.e.,
$y_j = \sum_iw_{ij}x_i$. The weight update rule that we use for the one-layer network is

\begin{equation}
\eqalign{
\Delta w_{ij} &= - \eta \frac{\partial L}{\partial w_{ij}}\cr
&= \eta \rho'(y_j)x_i(d_j-\rho(y_j)).
}
\label{equa:updateRule_1layer}
\end{equation}
 $\eta$ is the learning rate, and $\rho'(y_j)$ is the derivative of activation $\rho$.

We compare our learning rule with WTA based Hebbian learning, which introduces competition among neurons, ensuring that each neuron learns to recognize different patterns. Its classic formulation is \citep{lagani2023synaptic}:
\begin{equation}
\eqalign{
&\Delta w_{ij} = \eta x_i r_j;\cr
r_j &= \left\lbrace
\eqalign{
1 \quad \mbox{if } y_j &= max(y),\cr
0 \quad \mbox{if } y_j &\neq max(y);\cr
}
\right.
}
\label{equa:Hebb_WTA}
\end{equation}
where $r_j$ denotes the activation state of neuron $j$ in the hidden layer following the WTA selection, and vector $y$ contains all the aggregated input values in the hidden layer. 
($x_i$ and $y_j$ retain the same definition as in Eq.~\ref{equa:updateRule_1layer}.)

Hebbian learning, therefore, interprets the WTA mechanism as a nonlinear activation function for hidden neurons. In contrast, our approach utilizes WTA to define an unsupervised target, maintaining the original activation function $\rho$. This WTA definition is suitable for end-to-end training, allowing the entire network to be trained simultaneously, whereas the previous WTA definition is appropriate for layer-wise training, which enables independent training of each layer. While the weight updates in both our method and the Hebbian rule are proportional to the input value $x$, our approach introduces negative weight updates for non-winning neurons. Additionally, as the output of winning neurons approaches the value '1', the magnitude of their positive weight updates diminishes. In contrast, Hebbian learning keeps the weights of non-winning neurons static, updating only those of winning neurons, solely based on the input value $x$. Our weight update strategy effectively prevents unbounded growth in synaptic weights, promoting a highly sparse output in the last layer's neurons. However, this approach also increases the disparity between winning and non-winning neurons, making the use of homeostasis regularization critical for effective pattern recognition. This form of regularization is less critical in traditional Hebbian learning, as discussed by \cite{miconi2021hebbian}.

\begin{table}[htbp]
\caption{\label{tab:unsupervised_relative_compare}Comparison of test accuracy across unsupervised methods on the MNIST dataset for MLP SNNs and ANNs. Methods are categorized based on their class association approach: direct association (top) and linear classifier (bottom). Labeled data is exclusively utilized during the class association phase.}
\begin{minipage}{\textwidth}
    \centering
    \resizebox{1\columnwidth}{!}{
    \begin{tabular}{ccccc}
    \toprule
     Association &Architecture & Learning rule & Labels used  & Accuracy \\
    \midrule
    &SNN: \; 784-100  \citep{Nessler2013} & STDP & not mentioned  & 80.1\% \\
    &SNN: \; 784-300 \citep{Querlioz2013} & STDP & 1,000  & 93.5\% \\
    &SNN: \; 784-6400 \citep{DiehlCook2015}  & STDP & 60,000  & 95.0\% \\
    &ANN: \; 784-2000 \cite{moraitis2022softhebb}  & Hard WTA & 60,000  & 96.2\% \\
    Direct&ANN: \;784-2000 \cite{moraitis2022softhebb}  & SoftHebb & 60,000  & 96.3\% \\
    Association &ANN: \; 784-2000 (Ours) & BP & 60,000 & 95.78($\pm$0.14)\%\\
    &ANN: \; 784-2000 (Ours) & EP & 60,000 & 95.77($\pm$0.08)\%\\
    &ANN: \; 784-2000-2000 (Ours) & BP & 60,000 & 96.61($\pm$0.07)\%\\
    &ANN: \; 784-2000-2000 (Ours) & EP & 60,000 & 96.60($\pm$0.15)\%\\
    \midrule
    &ANN: \; 784-2000-10 \citep{moraitis2022softhebb} & Hard WTA& 60,000  & 97.8\% \\
    & ANN: \; 784-2000-10 \citep{moraitis2022softhebb} & SoftHebb & 60,000  & 97.8\% \\
    &ANN: \; 784-2000-10 \citep{Krotov2019CompetHidden}) & Hebbian-like & 60,000  & 98.5\% \\
    Linear&ANN: \; 784-2000-10 (Ours)  & BP & 60,000 & 97.08($\pm$0.11)\%\\
     Classifier&ANN: \; 784-2000-10 (Ours)  & EP & 60,000 & 97.25($\pm$0.07)\%\\
    &ANN: \; 784-2000-2000-10 (Ours)  & BP & 60,000 & 97.60($\pm$0.11)\%\\
    &ANN: \; 784-2000-2000-10 (Ours)  & EP & 60,000 & 97.57($\pm$0.10)\%\\
    
   \bottomrule
\end{tabular}
}
\end{minipage}
\end{table}

In Table \ref{tab:unsupervised_relative_compare}, we compare the results obtained with our unsupervised approach on the MNIST dataset to several Hebbian-like learning algorithms. The table differentiates between methods that utilize direct association and those that employ a linear classifier. As discussed in Results, the use of linear classifiers helps to further improve the accuracy, but the direct association method gives a better appreciation of the intrinsic quality of unsupervised learning.

Our result on MNIST outperforms the Spike-timing-dependent plasticity (STDP), a Hebbian learning implementation on the spiking neural network (SNN). It also achieves close results to standard Hebbian learning (Hard WTA) and SoftHebb \cite{moraitis2022softhebb}. For the direct association method, our one-layer network is 0.5\% below SoftHebb, but our two-layer network reaches 96.6\%, which is 0.3\% above the SoftHebb result. Our best result with the linear classifier association method is 97.3\% with the one-layer network, and 97.60\% with the two-layer network, compared to 97.8\% for SoftHebb and 98.5\% for \cite{Krotov2019CompetHidden}.

SoftHebb, developed by \cite{moraitis2022softhebb}, combines Oja's rule \citep{oja1982simplified} with a soft WTA. Oja's rule with the form $\Delta w_{ij} = \eta y_j(x_i - y_jw_{ij})$ introduces a weight normalization mechanism to prevent the uncontrolled growth of synaptic weights in classical Hebbian learning. \cite{moraitis2022softhebb} introduce WTA into Oja's rule inside the activation function of aggregated input $y_j$:
\begin{equation}
\eqalign{
    &\Delta w_{ij} = \eta r_j(x_i - y_jw_{ij});\cr
    &r_j = \mbox{softmax}(y_j) = \frac{e^{y_j}}{\sum_{k}e^{y_k}}.
}
\label{equa:Softhebb}
\end{equation}

Compared to the Hard-WTA implementation described before, softmax activation brings smoother weight changes for Hebbian learning, as all neurons can update their weights proportionally to their normalized exponential value, rather than the sparse weight update of the Hard-WTA version. Contrary to our method, neither Soft-WTA nor Hard-WTA includes negative weight updates for non-wining neurons (which is known as the anti-Hebbian part).

Our algorithm cannot directly benefit from the use of a soft-WTA: The Hard-WTA in our target definition ensures that the winner's weights are strengthened, as the output neurons value is bounded between 0 and 1. Using a softmax to define the target drives the output neurons to smoother and similar values, making them indistinguishable. However, we have introduced in our simulations a Label Smoothing strategy \citep{labelsmooth2017} (see Appendices) that resembles the soft WTA mechanism, by setting the winner values to be slightly less than '1' and the values for non-winning neurons slightly higher than '0', to penalize an overconfidence in the unsupervised target decision. This method does help to train two-layer networks with EP, but we have not obtained any improvement for training with BP. This may be due to the different network architectures used for EP and BP. Indeed, in the bidirectional recurrent networks trained by EP, the output value and therefore the target value directly influences the hidden layer, even before the weight update.

\cite{Krotov2019CompetHidden} proposes a Hebbian-like plasticity, which performs very well for the one-layer unsupervised learning case. They generalize the $L^2$ norm weight normalization in Oja's rule to the $L^p$ norm, and add an anti-Hebbian part for the least active neurons:
\begin{equation}
\eqalign{
r_j &= \left\lbrace
\eqalign{
1 \quad &\mbox{if } j=K,\\
-\Delta \quad &\mbox{if } j=K-k;\\
0 \quad &\mbox{otherwise. } 
}
\right.
}
\label{equa:Krotov2019}
\end{equation}
In the notations of \cite{Krotov2019CompetHidden}, $j$ ranks the neurons, with $j = 0$ corresponding to the least-activated, and $j=K$ the strongest-driven hidden unit. For consistency with our notations, we rename the competitive learning mechanism applied to their activation function as $r$.

\cite{Krotov2019CompetHidden} keeps the hard K-WTA as the competition mechanism, but adds negative weight changes to the $k$ neurons ranked below the winner. They also use the Rectified Power (RePU) activation function and gain about 6\% accuracy by optimizing the power coefficient. While such activation function cannot be directly combined with our unsupervised training method, it is possible that other choices of bounded activation functions at the output layer can increase the overall accuracy.

The Hebbian-like plasticity mechanism introduced by \cite{Krotov2019CompetHidden} surpasses our one-layer network by $1.2\%$. We show in the next section that adding a hidden layer to our network improves the quality of feature representation, underscoring the advantages of end-to-end training in unsupervised learning contexts. Furthermore, Hebbian learning struggles to integrate with supervised learning to fully exploit label information. The integration of Hebbian learning within a semi-supervised framework will be further discussed in the Discussion section \ref{section:semi_compare}.


\subsection{Two-layer Unsupervised Network: Comparison with Supervised End-to-End Training}

In the Results section, we show that by adding a hidden layer, we achieve better accuracy and clearer learned features at the output layer. This improvement demonstrates the advantage of end-to-end training on the multi-layer network, even for unsupervised training. In this part, we analyze the role of hidden layer in our unsupervised training method by comparing our results with supervised end-to-end training.

\begin{figure}[htbp]
    \centering
    \includegraphics[width=1\textwidth]{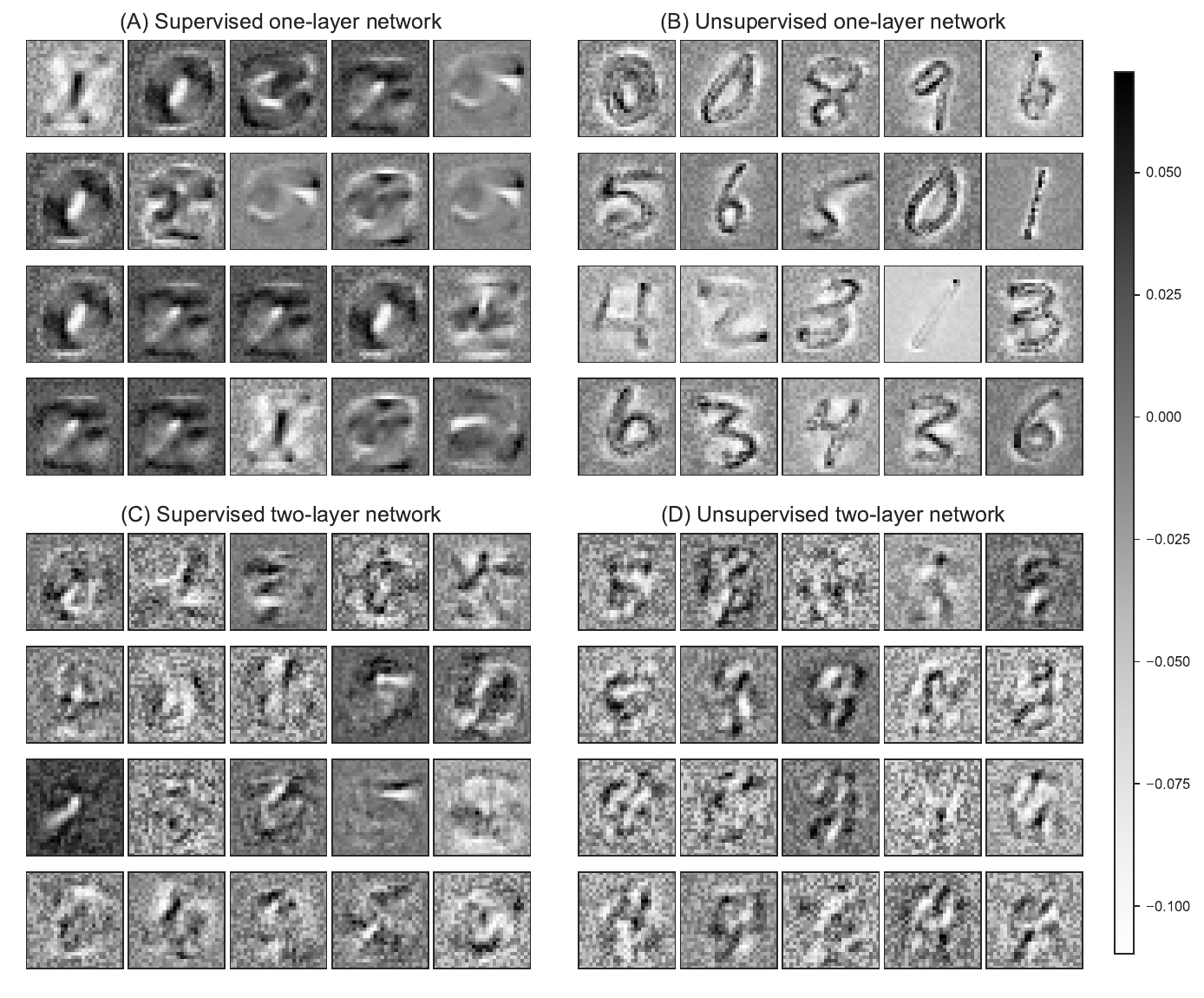}
    \caption{Visualization of the weights trained by BP on the MNIST dataset for 20 neurons in the first layer of (A) a supervised one-layer network, (B) an unsupervised one-layer network, (C) a supervised two-layer network and (D) an unsupervised two-layer network. The one-layer network comprises 2,000 output neurons, while the two-layer network includes an additional 2,000 hidden neurons.}
    \label{fig:weights_visu}
\end{figure}

We use BP to train supervisedly a one-layer and a two-layer network with the same architecture as used for unsupervised learning: 2,000 output neurons (and 2,000 hidden neurons for two-layer network). In this structure, each class is represented by 200 output neurons, that all have the same target '0' or '1' depending on the presented image. 

Figures \ref{fig:weights_visu}(A) and (B) illustrate the weight patterns of 20 randomly selected neurons from the output of a single layer network. In the supervised training scenario, depicted in Figure \ref{fig:weights_visu}(A), each neuron's blurred digit representations indicate that neurons learn to recognize all images within a class. Conversely, Figure \ref{fig:weights_visu}(B) demonstrates that in unsupervised training, neurons develop sharp, distinct patterns for individual handwritten digits within a class. Compared to supervised training, unsupervised training results in neurons learning varied weights even when they represent the same class. This variation is due to the selection of only $k$ strongest winners for each image in unsupervised training, enabling neurons to learn different images despite representing the same class.

We then examine in Figure \ref{fig:weights_visu}(C) and (D) the learned weights in the first layer of a two-layer network. Supervised end-to-end training allows hidden layers to establish a hierarchy of features, with lower layers capturing simple elements such as edges \citep{Erhan2009VisualizingHF}. In Figure \ref{fig:weights_visu}(C), the supervised training of the hidden layer reveals weights with significantly high or low values at the edges, facilitating the abstraction of simpler, more analyzable features for subsequent layers. Similarly, the hidden layer weights in our unsupervised network exhibit patterns reminiscent of digit components, including notably high values at edges. However, as shown in Figure \ref{fig:weights_visu}(D), these weights display more uniform patterns and less specificity across different edges than for supervised learning. 
This observation can explain why the improvement in accuracy introduced by the hidden layer is lower in our unsupervised learning than in supervised end-to-end training. Incorporating regularization techniques, such as homeostasis, into the hidden layer could improve both accuracy and feature extraction capabilities.

\subsection{Unsupervised CNN Network: comparison with other purely unsupervised learning}
\label{sec:cnn_compare}

\begin{table}[htbp]
\caption{\label{tab:cnn_compare}Comparison of test accuracy across unsupervised methods with a convolutional neural network on different datasets: MNIST, Fashion-MNIST, and SVHN. All labeled data is used during the class association phase.}
\begin{minipage}{\textwidth}
    \centering
    \resizebox{1\columnwidth}{!}{
    \begin{tabular}{ccccc}
    \toprule
     Dataset & Method & Architecture & Association  & Accuracy \\
    \midrule
     & Unsupervised BP (Ours) & 32f5-128f3-3000 & Linear classifier  & 99.17($\pm$ 0.05)\% \\
     & PL-AE \citep{bouayed2020pseudo} & 32f4-64f4-128f4-256f4-350 &  Linear classifier & 99.3\% \\
     & WTA-based \citep{similarity2023erkoc} & 97f5-117f5 & MLP & 99.18\% \\
     MNIST & Hard WTA \citep{lagani2022comparing} & 96f5-128f3-192f3 &  Linear classifier & 98.6\% \\
     & SoftHebb \citep{journe2022hebbian} & 96f5-384f3-1536f3 &  Linear classifier & 99.35\% \\
     & AE \citep{cavallari2018unsupervised} & 4f5-392-392-128 & SVM & 94\% \\
    \midrule
     & Unsupervised BP (Ours) & 32f5-128f3-3000 & Linear classifier &  90.25($\pm$ 0.25)\% \\
     & WTA-based \citep{similarity2023erkoc} & 92f3-48f3 & MLP & 90.11\% \\
    Fashion-MNIST  & AE \citep{cavallari2018unsupervised} & 4f5-392-392-128 & SVM & 88\% \\
    & Siamese Network\citep{trosten2019unsupervised} & 32f5-32f5-64 & SVM & 86\% \\
    \midrule
     & Unsupervised BP (Ours) & 32f5-128f3-3000 & Linear classifier & 81.53($\pm$1.15)\%\\
     & VAE \citep{pihlgren2020improving} &  32f4-64f4-128f4-256f4-128 & MLP & 81.2\%\\ 
     SVHN & VAE \citep{pihlgren2020improving} & 32f4-64f4-128f4-256f4-128 &  Linear classifier & 71.2\% \\
     & PL-AE \citep{bouayed2020pseudo} & 32f4-64f4-128f4-256f4-350 &  Linear classifier & 76.5\% \\
     & Siamese Network\citep{trosten2019unsupervised} & 32f5-32f5-64 & SVM & 78\%\\
   \bottomrule
\end{tabular}
}
\end{minipage}
\end{table}

Here, we demonstrate that our method can be generalized to CNNs and achieves high accuracy compared to other purely unsupervised methods, showing the effectiveness of unsupervised learning with a Self-Defined Target.

Table \ref{tab:cnn_compare} compares our test accuracy reached on MNIST, Fashion-MNIST, and SVHN with the results obtained through other purely unsupervised methods. These methods include reconstructive strategies like Autoencoders (AE) and Variational Autoencoders (VAE), competitive learning techniques such as Competitive WTA and Hebbian learning, and similarity-based models like Siamese networks. All these approaches utilize the classification task as the benchmark for evaluating their training outcomes. As for the network architecture, the number of channels and filter sizes define each convolutional layer; for instance, '32f5' signifies 32 output channels with a filter size of 5. A fully connected layer, like '3000', is characterized solely by its neuron count, indicating 3000 neurons.

While the pseudo-labelling auto-encoder (PL-AE) method \citep{bouayed2020pseudo} and the SoftHebb method surpass our results on MNIST (99.3\% and 99.35\%, respectively, compared to our 99.17\%), their network structures are more complex, with four and three convolutional layers, respectively, compared to two in our case. Furthermore, the SoftHebb method includes batch normalization before each convolutional layer, which complicates the calculation.

Conversely, for the Fashion-MNIST and SVHN datasets, we obtain better accuracy than PL-AE \citep{bouayed2020pseudo}, AE \citep{cavallari2018unsupervised}, VAE \citep{pihlgren2020improving}, Hard WTA \citep{lagani2022comparing}, and Siamese Network\cite{trosten2019unsupervised}, demonstrating the effectiveness of our method for unsupervised CNN training.

For more complex datasets like CIFAR-10 \citep{krizhevsky2009cifar}, further exploration of network architectures and normalization techniques is necessary to achieve better performance \citep{he2016deep, ioffe2015batch, miconi2021hebbian, journe2022hebbian}. Our preliminary results on the CIFAR-10 dataset achieved a test accuracy of $71.5(\pm 0.3)\%$ with a 96f5-384f3-5000 architecture. We believe that optimizing the network architecture and integrating advanced normalization techniques could further enhance accuracy in future work.


\subsection{Semi-supervised Approach for Neuromorphic Implementations}
\label{section:semi_compare}

Semi-supervised learning is ideally suited for neuromorphic edge computing environments, characterized by limited labeled data and a wealth of unlabeled data. As demonstrated in the Results section, our algorithm is well suited to this context, leveraging the available labeled data and benefiting from the generalization capabilities derived from unsupervised learning. This section compares our method with alternative semi-supervised strategies within the context of edge neuromorphic computing.

Layer-wise, bio-inspired algorithms like STDP and Hebbian learning, which traditionally lack feedback mechanisms, encounter challenges in implementing efficient supervised learning through solely forward processes. To address this, the concept of weakly supervised learning has been introduced, incorporating feedback limited to the network's last layer \citep{gupta2022bio, mozafari2018bio}. In this approach, the backward error, defined as the difference between the network's prediction and the actual label, is utilized to adjust synaptic strengths of the last layer. \cite{Lee2023semi} combine this weakly supervised STDP with unsupervised STDP to realize semi-supervised learning in a single-layer network for semi-supervised learning, outperforming purely unsupervised learning. Using 30\% of the labels, they achieve a 94.95\% test accuracy on  MNIST. Our approach appears more powerful in this context, as our semi-supervised EP training reaches 96.18($\pm$0.57)\%  test accuracy with only 0.5\% of the labels (Figure \ref{fig:Semi_BPEP}). Furthermore, \cite{Lee2023semi} obtained less than 60\% test accuracy when only weakly supervised STDP is used in the purely supervised context, indicating a limited efficiency compared to backpropagation, particularly with plentiful labeled data.

Another promising avenue explored in semi-supervised learning combines forward bio-inspired algorithms with end-to-end training through backpropagation \citep{lagani2021evaluating, furuya2021semi}. This hybrid method markedly improves the performance of bio-inspired algorithms in multi-layer fully connected networks. However, applying this approach to on-chip semi-supervised training in neuromorphic hardware presents challenges. The distinct nature of supervised and unsupervised phases necessitates separate circuits, significantly complicating chip design and impacting compactness. In our proposed approach, the neural network circuit maintains a consistent configuration across both supervised and unsupervised phases. The sole modification occurs in the small circuit segment responsible for computing the target, which is determined by either the label or autonomously generated at the network's output.

End-to-end training therefore offers a more cohesive approach to semi-supervised learning compared to layer-wise training, given its capacity to define a global loss in both supervised and unsupervised contexts. The method benefits from the precision of supervised learning through gradient descent weight updates. By maintaining a consistent global loss formula for both types of training and only substituting the supervised target with an unsupervised one in the absence of labels, we draw inspiration from the Pseudo Label (PL) method introduced by \citep{Lee2013}. The PL method starts with supervised training using a limited number of labels and then employs the pre-trained network's predictions as pseudo labels for further unsupervised training. Similarly, our approach defines these pseudo labels, comparable to our unsupervised target, through Winner-Take-All (WTA) selectivity, especially when the winner count, $k$, is set to one. Moreover, we incorporate the homeostasis mechanism into our target definition, enhancing semi-supervised learning performance.

\begin{table}[htbp]
\caption{\label{tab:semi-supervised}Test accuracy (\%) of semi-supervised methods over MNIST dataset.}
\begin{minipage}{\textwidth}
    \centering
    \resizebox{1\columnwidth}{!}{
    \begin{tabular}{ccccccccc}
    \toprule
     Learning method & 100 labels & 600 labels & 1000 labels & 3000 labels \\
     \midrule
     supervised BP + PL \citep{Lee2013} & 83.85\%   & 94.97\%  &  95.70\%  & 97.20\%\\
     \midrule
    supervised BP + WTA + homeostasis (Ours) & 88.43($\pm$1.60)\% &  94.88($\pm$0.35)\%  & 95.66($\pm$0.27)\%  & 96.76($\pm$0.11)\%  \\
    supervised EP + WTA + homeostasis (Ours) & 88.41($\pm$3.13)\% &  96.57($\pm$0.22)\%  & 96.79($\pm$0.12)\%  & 97.03($\pm$0.17)\%  
     \\
   \bottomrule
\end{tabular}
}
\end{minipage}
\centering
\end{table}

In Table \ref{tab:semi-supervised} we compare our semi-supervised learning results with those from the Pseudo Label (PL) method. Our approach exhibits a notable advantage when the count of labeled data is less than 1000. For example, with 600 labels, our semi-supervised  method achieves approximately 96.6\% test accuracy with Equilibrium Propagation (EP) training, compared to 95\% for PL. This gain largely comes from the homeostasis regularization incorporated in the unsupervised target decision, which is particularly effective in scenarios with very few labels. This observation aligns with the indispensable role of homeostasis during our unsupervised training. However, with more than 3,000 labeled data, PL performance exceeds that of our WTA by a margin of 0.4\%, achieving an error rate of 2.8\%. This performance gap can be linked to differences in the initial supervised training phase, where \cite{Lee2013} surpassed our approach by about 0.8\%. While we employ the Adam Optimizer with an exponential learning rate scheduler, \cite{Lee2013} uses a Stochastic gradient descent optimizer combined with an exponentially decaying learning rate and momentum term. This weight adjustment strategy seems to offer an advantage when training in a supervised manner with sparse data.

\cite{Lee2013} also proposes to integrate layer-wise unsupervised training techniques, such as the Denoising Auto-Encoder (DAE), before the first stage of supervised training. In scenarios with fewer labels, DAE is effective, boosting results by 5.7\% and 1\% for 100 and 600 labels case, respectively. Such layer-wise pre-training techniques could complement our method in the future to improve performance, but require considerable circuit overhead for edge computing. 

Emerging semi-supervised training methods, inspired by the pseudo label concept, have shown significant promise in complex class classification tasks within the machine learning (ML) domain \citep{laine2016temporal, tarvainen2017mean, cascante2021curriculum}. The approach of \citep{asano2019self} which defines a target using network dynamics, closely aligns with our method. These innovative ML strategies hold potential for enhancing our method's effectiveness. However, their deployment demands extensive on-chip memory and peripheral circuits, rendering them less suitable for low-power, compact edge AI systems, unlike our approach. 


\section{Conclusion}

In this study, we use bio-inspired mechanisms to create a self-defined target, serving as a simple drop-in replacement for standard loss functions during unsupervised learning phases. This method is suitable for both semi-supervised learning and unsupervised end-to-end training. Being training-algorithm and architecture agnostic, it holds promise for rapid integration into various existing hardware platforms.

We show competitive results on MNIST with SOTA approaches in unsupervised and in semi-supervised learning with both global (BP) and local learning rules (EP). Our approach also benefits from the inclusion of deeper fully connected layers. Furthermore, we show that our method achieves high accuracy for CNNs applied to MNIST, Fashion-MNIST and SVHN datasets.
These results represent a step forward toward edge AI hardware that can learn by leveraging both labeled and unlabeled data.


\section{Appendices}

\subsection{WTA selectivity}

The Winner-Take-All (WTA) mechanism is widely used in unsupervised learning \citep{Ferre2018WTA, oster_computation_2009}. Within a network layer, neurons compete against one another, with only the neuron exhibiting the highest activation allowed to reinforce its synaptic connections. In our study, WTA selectivity is employed to obtain the unsupervised target, which helps to further calculate the unsupervised loss.

The unsupervised target for a strict WTA is defined as $\boldsymbol{d}$, where $d_i$ is the $i^{th}$ element of $\boldsymbol{d}$, and $\boldsymbol{y}$ is the output of the network:
\begin{equation}
    d_i = \left\lbrace
    \eqalign{
        1 \quad &\mbox{ if } y_i = max(\boldsymbol{y}),\cr
        0 \quad &\mbox{ if } y_i \neq max(\boldsymbol{y}).
}
     \right.
\end{equation} 

This approach implicitly necessitates scaling the output values between 0 and 1, assigning the winning neuron a target of maximum activation value ('1') and all other neurons a target of minimum activation value ('0'). 

For a large number of output neurons, selecting a single winner among all the output neurons is not the optimal choice. We thus use the k-winners-take-all (kWTA) method \citep{Majani1988}:

\begin{equation}
    d_i = \left\lbrace
    \eqalign{
        1 \quad  y_i \in \{ k \mbox{ largest  elements of } \boldsymbol{y} \},\cr
        0 \quad \mbox{Otherwise}.
       }
     \right.
\end{equation}

During the simulation, $k$ is treated as a hyperparameter, which is found to be strongly dependent on the number of output neurons and the dropout probability for the output layer. Specifically, a higher number of output neurons leads to an increased selection of winners $k$.

However, relying solely on WTA selectivity for determining unsupervised targets proves to be ineffective. Minimizing the difference between the output value and the unsupervised target determined solely by WTA leads to a scenario where the initial 'winner' neuron ends up learning all data classes, undermining the intended goal of achieving distinct classification. This issue arises because the weights of the 'winner' neuron are strengthened after processing initial images, while the weights of other neurons are weakened. As a result, the other non-winning neurons become increasingly disadvantaged, and in some cases, may even become silent during subsequent learning phases.

\subsection{Homeostasis}

Homeostasis is a bio-inspired regulation method often employed in STDP algorithms, as referenced by \citep{DiehlCook2015, Querlioz2013}. At its essence, homeostasis aims to mitigate excessive learning by individual neurons. Within the realm of SNNs, this equilibrium is maintained by modulating the neurons' threshold levels, as done in \citep{Querlioz2013}; here, the adjustment is critical since a neuron's learning is contingent upon its ability to spike. In contrast, ANNs function based on continuous values rather than discrete spiking events. Consequently, in ANNs—and particularly in the context of end-to-end training—the mere magnitude of a neuron's activation does not dictate the occurrence of learning. Instead, learning is determined by whether a neuron's activation outcompetes that of its peers. Therefore, in adapting the concept of neural activity from SNNs to ANNs, it is the frequency with which a neuron 'wins' this competitive process that should be regulated. This approach ensures a balanced distribution of learning opportunities across the network, effectively mirroring the principle of homeostasis.

In our approach, the homeostasis term is applied to each output neuron; if a particular output neuron is consistently selected as the 'winner' too many times, we increment its homeostasis term $\boldsymbol{H}$; otherwise, we decrement the term $\boldsymbol{H}$. The update to the homeostasis term is performed after each weight update, following the rule outlined below:
\begin{equation}
    \Delta H = \gamma (\boldsymbol{A} - \boldsymbol{T}).
    \label{equa:Homeo}
\end{equation}
Here, $A$ is the mean activity, $T$ is the target activity, and $\gamma$ is a multiplicative constant.

\begin{figure}[htbp]
    \centering
    \includegraphics[width=1\textwidth]{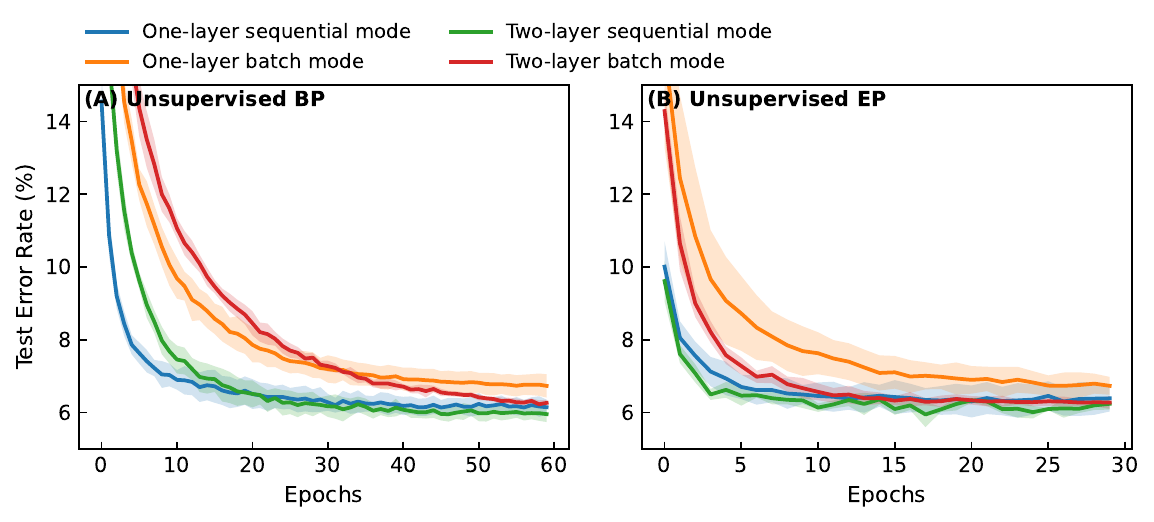}
    \caption{\label{fig:mode_compare} MNIST test error rate (\%) comparison between batch and sequential mode for different networks trained with \textbf{(A)} unsupervised BP and \textbf{(B)} unsupervised EP. The one-layer network has 500 output neurons. A hidden layer with 1024 neurons is added in the two-layer unsupervised network.}
    
\end{figure}

The target activity $\boldsymbol{T}$ is defined by a fixed value $\frac{k}{N_{output}}$ for all the output neurons. The mean activity $\boldsymbol{A}$ is calculated as a moving average of the unsupervised target. In the following, $<\boldsymbol{d}>_P$ represents the moving average of target $\boldsymbol{d}$ over $P$ examples.
We considered two types of moving average:
\begin{itemize}
    \item Exponential moving average : 
    \begin{equation}
        \boldsymbol{A} = <\boldsymbol{d}>_P = \eta\boldsymbol{d} + (1-\eta)<\boldsymbol{d}>_{P-1}.
    \label{equa:EMA}
    \end{equation}
    In the exponential moving average (EMA), we take the average of all previously presented examples, which means that $P$ represents the total number of presented images. In simulations employing EMA, this approach is referred to as the sequential mode, as the input data is provided one example at a time.
    \item Classical moving average:
    \begin{equation}
        \boldsymbol{A} = <\boldsymbol{d}>_P = \frac{\sum _P \boldsymbol{d}}{P}.
    \label{equa:CMA}
    \end{equation}
    In the classical moving average (CMA), we take the average of $P$ examples, where $P$ is a fixed number. This moving average calculation can be easily implemented in the mini-batch mode computation and is referred as batch mode in the following.
\end{itemize}
In the simulations, both types of moving averages—exponential and classical—exhibit similar performances (Figure \ref{fig:mode_compare}), sequential mode using the EMA has a slightly better performance than the batch mode using the CMA. We opt for the classical moving average to test our algorithm because the mini-batch mode facilitates parallelization of calculations. In real-world applications, the choice of moving average can be customized to suit the specific requirements of practical problems.

The hyperparameter $\gamma$ and $\eta$ do not have a large influence on the accuracy. For both EP and BP optimized $\gamma$ is about 0.5.

\subsection{Influence of Output Neuron Count}

The effectiveness of unsupervised learning, facilitated by WTA selectivity, is critically dependent on the number of output neurons participating in the competition \citep{Querlioz2013, DiehlCook2015}. Here, we examine how the count of output neurons impacts classification accuracy. 

\begin{figure}
    \centering
    \includegraphics[width=1\textwidth]{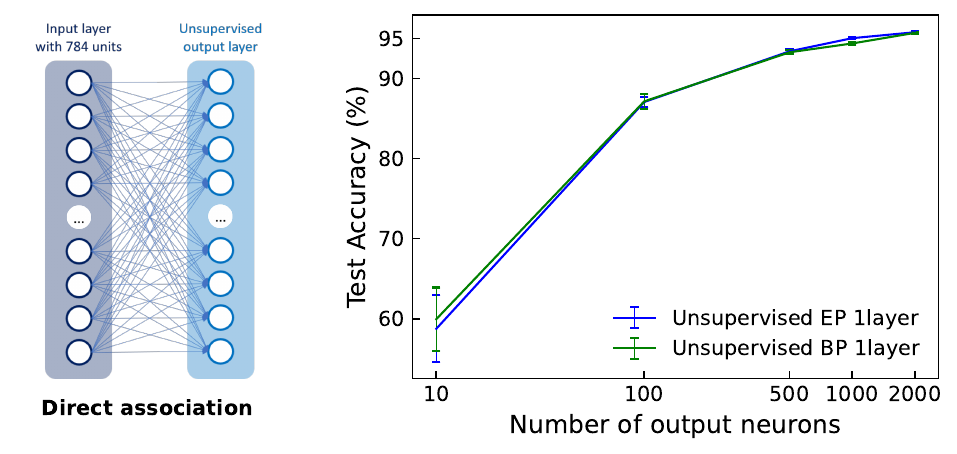}
\caption{\label{fig:output_effect_1layer} Performance comparison of a single unsupervised layer using different numbers of neurons at the output, trained either with EP or with BP. Following unsupervised training, a direct association is performed with 2\% of the training data. \textbf{Left:} Structure of the single-layer network; \textbf{Right:} Test accuracy on the MNIST dataset as a function of the number of neurons in the output layer.}
\end{figure}

The graph in Figure \ref{fig:output_effect_1layer} demonstrates that the unsupervised learning performance is significantly influenced by the number of output neurons. Specifically, when the number of output neurons is limited to 10, the test accuracy achieved is approximately 60\%.  This is attributed to inadequate class representation, where not all classes are distinctly represented by the available neurons. As the number of output neurons increases, each class is more likely to be uniquely represented, enhancing classification accuracy. For example, with 500 output neurons, a test accuracy of 93.4\% for unsupervised EP and 93.3\% for unsupervised BP is achieved, which is only 2.4\% and 2.5\% lower, respectively, than the highest accuracy obtained with 2000 output neurons.

\subsection{Dropout Regularization}

In our study, we employed dropout regularization — a technique traditionally utilized in machine learning to counteract overfitting \citep{Dropout_initial} — and observed a marginal enhancement in model performance. We treated the dropout probability as a hyperparameter, optimizing its value through grid research between 0, 0.1, 0.2, 0.3, 0.4 and 0.5. 

For unsupervised training with BP and EP, the optimal dropout probabilities were approximately 0.2 for output neurons and zero for hidden neurons. These findings suggest that within the context of unsupervised learning, the primary role of dropout may be to balance neuron activity rather than to prevent overfitting. Following the methodology used by \cite{hinton2012dropout}, dropout is also employed on the input neurons with a fixed probability of 0.3.

For semi-supervised learning, dropout with probabilities of 0.3 and 0.5 was applied to the input and hidden neurons, respectively. Given that the number of output neurons corresponds to the number of classes, we did not apply dropout to the output layer in this case.

\subsection{Label Smoothing}

We use label smoothing  to penalize the overconfidence \citep{labelsmooth2017} of unsupervised targets :

\begin{equation}
    \boldsymbol{\tilde{d}} = \boldsymbol{d}*(1-\alpha) + \boldsymbol{T}*\alpha,
\end{equation}

We maintain the same notations as previously introduced: $\boldsymbol{d}$ represents the unsupervised target vector, and $\boldsymbol{T}$ denotes the target activity vector. Each element of $\boldsymbol{T}$ has the same value, given by $\frac{k}{N_{output}}$, for all output neurons. This technique moderates the tendency of output neurons to assume extreme values of '1' or '0'. Instead, it adjusts $k$ winner neurons towards a more moderated value, while diminishing the inhibition on other neurons. The sum of the adjusted unsupervised target $\tilde{\boldsymbol{d}}$ remains $k$. In our simulations, the smoothing factor $\alpha$ is set to 0.3.

\subsection{Hyperparameters}

In our study, the hyperparameters are searched using Random Search with the Optuna framework \citep{optuna_2019}.

\subsubsection{Unsupervised Learning}

 For unsupervised learning, mean squared error (MSE) is taken as the loss function, and the gradients are updated by stochastic gradient descent (SGD). 

 For \textbf{activation functions}:
 \begin{itemize}
     \item In unsupervised EP, hardsigmoid is taken as the activation function for all the layers. 
     \item In unsupervised BP, ReLu is the hidden layer activation function, and hardsigmoid is the output layer activation function.
 \end{itemize}
 
\begin{equation}
\eqalign{
\centering
    \mbox{relu}(x) &= max(0,x)\cr
    \mbox{hardsigm}(x) &= min(max(0,x), 1)
}
\end{equation}

We train separately a single unsupervised layer with 2,000 output neurons, and a two-layer unsupervised network with 2,000 added hidden neurons. Dropout regularization is applied to both the input and output layers with probabilities of 0.3 and 0.2, respectively. To improve convergence during unsupervised training, we employ a linear learning rate scheduler that starts with a ratio of 1 and decreases to 5e-4. Other hyperparameters are shown in Table \ref{tab:hyper_epbp}. 

\begin{table}[H]
    \caption{\label{tab:hyper_epbp} Hyperparameters of one-layer / two-layer network trained by unsupervised EP/BP.}
    \centering
    \begin{minipage}{0.9\textwidth}
    \resizebox{1\columnwidth}{!}{
    \begin{tabular}{ccccc}
    \toprule
     Hyperparameters & EP one-layer & EP two-layer & BP one-layer & BP two-layer \\
    \midrule
    T & 40 & 60 & - & -\\
    K & 10 & 20 & - & - \\
    $\beta$ & 0.2 & 0.2 & - & -\\
    Winners $k$ & 5 & 4 & 6 & 5 \\
    $\gamma$ & 0.3 & 0.65 & 0.4 & 0.5\\
    Batch Size & 16 & 32 & 16 & 16\\
    LR & 0.03  & 0.01-0.015 & 8 & 2.5-5\\
    Epoch & 100 & 100 & 200 & 200\\
    Label Smoothing& Yes & Yes & No & No\\
    \toprule
    \end{tabular}
    } 
    \end{minipage}
\end{table}

Following the unsupervised training, for the linear classifier association method, we append a supplementary linear classifier to the terminal layer of the unsupervised network, which enables the network to make predictions during inference. The input neuron count for this perceptron matches the output neuron count of the final unsupervised layer, while its output neuron count aligns with the number of classes.

To train the auxiliary linear classifier, we use Cross-Entropy loss for error computation and optimize it with the Adam algorithm. The hyperparameters required for training classifiers exhibit significant differences, depending on whether BP or EP is used as the unsupervised learning algorithm. The classifier is trained for 50 epochs when the unsupervised network is trained using BP and 100 epochs when trained using EP. A learning rate scheduler is employed during classifier training. For the BP-trained unsupervised network, the classifier learning rate undergoes exponential decay at each epoch with a decay factor of 0.9. Conversely, for the EP-trained unsupervised network, the classifier rate undergoes linear decay from a factor of 1 to 0.5 over the course of all training epochs.

The classifier learning rate for the different label numbers used for class association is detailed in Table \ref{tab:classifier_hyper}.

\begin{table}[H]
    \caption{\label{tab:classifier_hyper} Learning rate for the added linear classifier with different amount of available labels.}
    \begin{minipage}{\textwidth}
    \centering
    \resizebox{.9\columnwidth}{!}{
    \begin{tabular}{ccccc}
    \toprule
 & 1\% labels & 5\% labels & 50\% labels & 100\% labels \\
    \midrule
    BP unsupervisedly trained & 0.05 & 0.065 & 0.09 &  0.1 \\
    EP unsupervisedly trained & 0.0085 & 0.008 & 0.005 &  0.0022 \\
    \toprule
    \end{tabular}
    }
\end{minipage}
\end{table}

\subsubsection{Convolutional Layers}

\begin{table}[ht]
    \caption{\label{tab:cnn_archi} Convolutional network architecture used in both supervised and unsupervised learning.}
    \begin{minipage}{\textwidth}
    \centering
    \resizebox{.93\columnwidth}{!}{
    \begin{tabular}{cll}
    \toprule
    \#layer & MNIST/Fashion-MNIST/SVHN & CIFAR-10 \\
    \midrule
       &  5x5 conv32 padding=2 stride=1  & 5x5 conv96 padding=2 stride=1\\
    1  &  Relu & WTA\\
       &  4x4 Max/AvPool padding=1 stride=2 & 4x4 MaxPool padding=1 stride=2\\
    \midrule
      &  3x3 conv128 padding=1 stride=1 & 3x3 conv384 padding=1 stride=1\\
    2 &  Relu & WTA\\
      &  4x4 Max/AvPool padding=1 stride=2 & 4x4 MaxPool padding=1 stride=2\\
    \midrule
       & Dropout=0.2 & Dropout = 0.3\\
    3  & FC layer with 3000 neurons & FC layer with 5000 neurons\\
       & Dropout=0.3 & Dropout=0.5\\
    \toprule
    \end{tabular}
    }
\end{minipage}
\end{table}

The convolutional network architecture is shown in Table \ref{tab:cnn_archi}. No pre-processing or normalization was applied to any of the MNIST, Fashion-MNIST and SVHN. Random flip transformation is applied to the CIFAR-10 dataset.

For the MNIST dataset, we use average pooling after the convolutional layer, while for the other datasets, we use max pooling. The Adam optimizer is used to update the gradients, and no label smoothing is applied. For the CNNs training, the winner number $k$ is set at 1, and the homeostasis coefficient $\gamma$ is set at 1 as well. 

For CIFAR-10, we use a new activation function inspired by the Triangle method \citep{coates2011analysis}, referred to as `WTA' in the table. This activation function first subtracts the mean activation of each channel across all positions, rectifies negative values to 0, then subtracts the mean activation at each position across all channels, and rectifies negative values to 0 again. This procedure effectively performs two rounds of winner-take-all: first, across neurons within the same channel, and second, across neurons at the same position.

The weights of CNNs are initially pruned by 30\% for all datasets except CIFAR-10, where 80\% pruning is applied to introduce greater sparsity into the network, following the approach in \cite{miconi2021hebbian}. Dropout is applied both before and after the fully connected layer, which serves as the output layer in the CNN. All the other hyperparameters for CNN training can be found in Table \ref{tab:hyper_cnn}.

\begin{table}[H]
    \caption{\label{tab:hyper_cnn} Hyperparameters of CNNs trained by unsupervised BP.}
    \centering
    \begin{minipage}{0.93\textwidth}
    \resizebox{1\columnwidth}{!}{
    \begin{tabular}{ccccc}
    \toprule
     Hyperparameters & MNIST & Fashion-MNIST & SVHN & CIFAR10 \\
    \midrule
    Batch Size & 16 & 16 & 16 & 10 \\
    LR & (6e-7, 3e-8, 3e-7) & (5e-7, 3e-8, 3e-6) & (6e-9, 4e-9, 2e-7) & (3e-6, 7e-7, 8e-5) \\ 
    Epoch & 30 & 20 & 40 & 40 \\
    \bottomrule 
    \end{tabular}
    } 
    \end{minipage}
\end{table}

\subsubsection{Semi-supervised Learning}

In our semi-supervised learning approach, we employ the cross-entropy loss function to compute the output error. Gradient updates are performed using the Adam optimizer.

For \textbf{activation function}:
\begin{itemize}
    \item In the EP paradigm, we use the hardsigmoid function as the activation for all layers.
    \item In the BP paradigm, the hidden layer utilizes the ReLU activation function, while the sigmoid function is adopted for the output layer. 
\end{itemize}

Our training commences with $N_1$ epochs of supervised learning, using limited labeled datasets of sizes 100, 300, 600, 1000, and 3000, respectively, in four distinct scenarios. This is followed by an amalgamation of supervised and unsupervised training over the subsequent $N_2$ epochs.

The learning rate's evolution, especially its ratio between supervised and unsupervised learning, plays a pivotal role in determining the final model accuracy, as discussed in \cite{Lee2013}. To regulate this, we employ an exponential learning rate scheduler during the initial $N_1$ epochs of supervised pre-training. Subsequently, a linear learning rate scheduler is utilized for the following $N_2$ epochs, encompassing both supervised and unsupervised training phases.

\begin{itemize}
    \item During supervised pre-training, the network is trained using a limited set of labeled data for $N_1$ epochs. The initial learning rate for this phase is referred to as the Pre-training LR in Table \ref{tab:semi_hyper_epbp}. This learning rate undergoes exponential decay with a factor of 0.97 throughout the $N_1$ epoch.
    \item For the subsequent $N_2$ epochs of alternating supervised and unsupervised training, two distinct linear learning rate schedulers are applied separately for supervised and unsupervised training. The initial learning rate for these $N_2$ epochs is referred to as the Semi-supervised LR in Table \ref{tab:semi_hyper_epbp}. During supervised re-training, the learning rate begins at 0.7 times the Semi-supervised LR and gradually decreases to a factor of 0.05. Concurrently, for unsupervised training, the learning rate initiates at 0.001 times the Semi-supervised LR and concludes at a factor of 0.18.
\end{itemize}

The mini-batch size for supervised learning is 32, while for unsupervised learning is 256. Dropout is applied with 0.2 and 0.5 probabilities for the input layer and hidden layer separately. These are applicable to both EP and BP learning methods.

\begin{table}[H]
    \caption{\label{tab:semi_hyper_epbp} Hyperparameters of semi-supervised learning by EP/BP with different labels numbers. There is always 5,000 neurons in the hidden unsupervised layer, and 10 neurons in the output layer.}
    \begin{minipage}{\textwidth}
    \centering
    \resizebox{1\columnwidth}{!}{
    \begin{tabular}{lcccccc}
    \toprule
     &Hyperparameters & 100 labels & 300 labels & 600 labels & 1000 labels & 3000 labels \\
    \midrule
    &T & 60 & 60 & 60 & 60 & 60\\
    &K & 10 & 10 & 10 & 10 & 10\\
    &$\beta$ & 0.48 & 0.4 & 0.38 & 0.4 & 0.48 \\
    &$\gamma$ & 0.1 & 0.1 & 0.005 & 0.001 & 0.001\\
    Semi-supervised EP 
    &Epoch $N_1$ & 100 & 100 & 100 & 100 & 100\\
    &Epoch $N_2$ & 200 & 200 & 200 & 200 & 200 \\
    &Pre-training LR & 0.001-0.0035 & 0.001-0.0035 & 0.001-0.0035 & 0.001-0.0035 & 0.001-0.0035\\
    &Semi-supervised LR & 0.0003-0.009 & 0.0004-0.008 & 0.0002-0.001 & 0.0002-0.006 & 0.00015-0.009\\
    
    \midrule
    
    &$\gamma$ & 0.1 & 0.1 & 0.003 & 0.001 & 0.001\\
    &Epoch $N_1$ & 200 & 200 & 200 & 200 & 200\\
    &Epoch $N_2$ & 400 & 400 & 400 & 400 & 400\\
    Semi-supervised BP & Pre-training LR & 0.001-0.003& 0.001-0.003 & 0.001-0.003 &  0.001-0.003 & 0.001-0.003 \\
    &Semi-supervised LR & 0.0006-0.0044 & 0.0005-0.004 & 0.0001-0.002 & 0.00024-0.003 & 0.0006-0.0008 \\

    \toprule
    \end{tabular}
}
\end{minipage}
\end{table}

\section*{Conflict of Interest Statement}
The authors declare that the research was conducted in the absence of any commercial or financial relationships that could be construed as a potential conflict of interest.

\section*{Author Contributions}
The project was designed by JG and DL with the participation of DQ and JL. The code and numerical simulations were all carried out by DL. AP contributed to semi-supervised learning tests. DL, JG, JL and DQ participated to the writing of the manuscript and all the authors discussed and reviewed the content.

\section*{Funding}
This work was funded by the European Research
Council advanced grant GrenaDyn (reference: 101020684) and the ANR project SpinSpike ANR-20-CE24-0002-02. 

\section*{Acknowledgments}
This project was supported with AI computing and storage resources by GENCI at IDRIS thanks to Grant 2023-AD010913993 on the supercomputer Jean Zay's V100 partition. The text of the
article was partially edited by a large language model (OpenAI ChatGPT).
We greatly appreciate Dr. Xing Chen's constructive suggestions regarding the CIFAR-10 training process.

\section*{Supplemental Data}

The programming code utilized in this study has been made publicly available online under an open-source license: \url{https://github.com/neurophysics-cnrsthales/unsupervised-target}.

\bibliographystyle{agsm}
\bibliography{test}

\end{document}